\documentclass{hld2023}  %

\usepackage{algorithm}
\usepackage{algpseudocode}

\usepackage{amsmath}
\usepackage{float}
\usepackage{mathtools}
\usepackage{caption}
\usepackage{subcaption}
\usepackage{todonotes}

\DeclareMathOperator*{\argmax}{\arg\!\max}

\newcommand{\settingA}{{Offline RL: agent trained using experience generated from a greedy pre-trained policy}}

\newcommand{\settingB}{{Offline RL: agent trained using experience generated from an $\epsilon$-greedy pre-trained policy}}
\newcommand{\settingC}{{Offline RL: agent trained from the replay buffer of a pre-trained agent}}
\newcommand{\settingD}{{Online RL: agent trained from a replay buffer of its own experience}}

\firstpageno{1}

\begin{document}

\title[Investigating the Edge of Stability Phenomenon in Reinforcement Learning]{Investigating the Edge of Stability Phenomenon in Reinforcement Learning}
\hldauthor{%
\Name{Rares Iordan} \Email{rares-ioan.iordan.18@ucl.ac.uk}\\
\addr {University College London, United Kingdom} 
\AND
\Name{Marc Peter Deisenroth} \Email{m.deisenroth@ucl.ac.uk}\\
\addr {University College London, United Kingdom} 
\AND
\Name{Mihaela Rosca} 
\Email{mihaela.rosca.19@ucl.ac.uk}\\
\addr {University College London, United Kingdom} 
\thanks{Also at Google Deepmind.}
}

\maketitle

\begin{abstract}%
Recent progress has been made in understanding optimisation dynamics in neural networks trained with full-batch gradient descent with momentum with the uncovering of the edge of stability phenomenon in supervised learning \citep{edgeofstability}. 
The edge of stability phenomenon occurs as the leading eigenvalue of the Hessian reaches the divergence threshold of the underlying optimisation algorithm for a quadratic loss, after which it starts oscillating around the threshold, and the loss starts to exhibit local instability but decreases over long time frames.
In this work, we explore the edge of stability phenomenon in reinforcement learning (RL), specifically off-policy Q-learning algorithms across a variety of data regimes, from offline to online RL. Our experiments reveal that, despite significant differences to supervised learning, such as non-stationarity of the data distribution and the use of bootstrapping, the edge of stability phenomenon can be present in off-policy deep RL. Unlike supervised learning, however, we observe strong differences depending on the underlying loss, with  DQN --- using a Huber loss ---  showing a strong edge of stability effect that we do not observe with C51 --- using a cross entropy loss. Our results suggest that, while neural network structure can lead to optimisation dynamics that transfer between problem domains, certain aspects of deep RL optimisation can differentiate it from domains such as supervised learning.
\end{abstract}

\section{The Edge of Stability Phenomenon}

\citet{edgeofstability}  use a thorough experimental study to shed light on deep learning optimisation dynamics by showing that full-batch gradient descent training in supervised learning exhibits two phases: \textit{progressive sharpening} and \textit{edge of stability}. In the first stage of training, \textit{progressive sharpening}, the leading eigenvalue of the Hessian, $\lambda_1$, increases steadily and the loss decreases monotonically. As $\lambda_1$ increases, it reaches the  divergence threshold  of the underlying optimisation algorithm under a quadratic loss assumption (we will call this ``the quadratic divergence threshold");  for gradient descent with learning rate $\eta$ and momentum decay rate $\beta$, the quadratic divergence threshold is $\frac{1}{\eta} (2 + 2\beta)$. As $\lambda_1$ reaches the quadratic divergence threshold, \textit{the edge of stability phenomenon} occurs: the loss starts to exhibit short-term instabilities, while decreasing over long time scales; $\lambda_1$ no longer steadily increases, but fluctuates around the threshold. For cross entropy losses, the edge of stability area is followed by a  decrease in $\lambda_1$, while for mean squared error losses $\lambda_1$ stays in the edge of stability area. Similar results are shown for stochastic gradient descent, across batch sizes, though the effect is less pronounced as the batch sizes decrease \citep{edgeofstability}.

The edge of stability phenomenon shows that neural network training is strongly affected by the quadratic divergence threshold of the underlying optimisation algorithm, and exceeding it leads to training instabilities. This observation
has garnered a lot of  interest, with recent works having analysed the edge of stability phenomenon in supervised learning with both theoretical and empirical tools \citep{ahn2022understanding,ma2022quadratic,rosca2023continuous,chen2022gradient,damian2022self}. To the best of our knowledge, no studies on the edge of stability phenomenon have been made outside of supervised learning. We complement this body of work by empirically investigating whether the edge of stability phenomenon in occurs in off-policy deep RL algorithms DQN and C51 across a variety of data regimes, from offline to online learning. Upon acceptance, we will make the code and data used publicly available.

\section{Challenges with optimisation in off-policy deep reinforcement learning}
\label{sec:challenges_rl}

To investigate whether the edge of stability phenomenon translates to deep RL, we conduct experiments using DQN \citep{mnih2013} and C51 \citep{c51}, two off-policy algorithms that model the state-action value function $Q(s, a; \boldsymbol{\theta})$ with a neural network with parameters $ \boldsymbol{\theta}$. We study both DQN and C51 as their losses correspond to the mean squared error and cross entropy loss used in supervised learning, studied by \citet{edgeofstability} when investigating the edge of stability phenomenon. For DQN, we investigate the more commonly used the Huber loss, which is quadratic around 0:
\begin{equation}
    \min_{\boldsymbol{\theta}} E(\boldsymbol{\theta}) = 
\begin{dcases}
    \mathbb{E}_{(s, a, s', r) \sim \mathcal{R}} \  \tfrac{1}{2} \left(Q(s, a; \boldsymbol{\theta}) - \left(r + \gamma\max_{a'} Q(s', a'; \boldsymbol{\theta}) \right ) \right )^2 , & \text{if } (\cdot)^2 \leq 1 \\
    \mathbb{E}_{(s, a, s', r) \sim \mathcal{R}} \left|Q(s, a; \boldsymbol{\theta}) - \left(r + \gamma\max_{a'} Q(s', a'; \boldsymbol{\theta}) \right ) \right | - \tfrac{1}{2} , & \text{otherwise.}
\end{dcases}
\label{eq:dqn_huber}
\end{equation}
C51 \citep{c51}, the distributional counterpart of DQN, uses  distributional quantiles instead of operating in expectation as in Eq~\eqref{eq:dqn_huber}, leading to a cross entropy loss; for details, we refer to Appendix~\ref{app:algo}.

Off-policy deep RL algorithms like DQN and C51 differ from supervised learning both through their objectives---which use bootstrapping---and the data  present in the replay buffer $ \mathcal{R}$ used for learning  the agent, which can be non-stationary, have noise inserted to help exploration, and can be obtained from the agent's own experience. All the above affect optimisation dynamics in deep RL, and have been studied individually \cite{mnih2013, overcomenonstationarity, optimistic_offline_rl, noisyexploration, replaybufferinteraction}. Bootstrapping --- the dependence of the regression target in Eq \eqref{eq:dqn_huber} on the Q-function ---  can lead to increased variance and bias in model updates \cite{schaul2016prioritized, maei2011gradient}. To mitigate instabilities introduced by bootstrapping often a target network is used, where old parameters updated at regular intervals are used to construct the target instead of the current parameters; both DQN and C51 use target networks.
In online RL, where the replay buffer $\mathcal{R}$ is filled with the agent's own experience, the non-stationarity of the data present in the replay buffer $\mathcal{R}$ violates the i.i.d. assumption required by many optimisation algorithms and gradient updates might not form a gradient vector field \cite{bengio2020interference}. 
Offline RL \cite{offline_rl_no_exploration, offlinerldatasets, offlinerlstochasticpolicies}, where the agent learns from a fixed replay buffer often gathered from another agent or expert demonstrations, can mitigate some of the training challenges with online RL, but can suffer from poor agent performance due to distributional shift \cite{offlinerl} --- the discrepancy between the data distribution used for learning, present in the offline dataset (the replay buffer), and the distribution the policy encounters during execution. 

Given these peculiarities of RL, it is unclear how optimisation effects observed in supervised learning, such as the edge of stability results, transfer to deep RL, and they interact with the behaviour of the loss function. Since the value of the loss function in RL has not been connected with agent performance, many RL works do not study or report loss function behaviour, and focus on the agent reward instead. Here, we focus on the behaviour of the loss function in deep RL and aim to connect it with the value of the leading Hessian eigenvalue through edge of stability results.

\section{Investigating the edge of stability in phenomenon reinforcement learning}

When studying the edge of stability phenomenon in deep RL,
we aim to isolate the effect of the data distribution from the other aspects of RL, such as the use of bootstrapping. We thus train agents across multiple data regimes, ranging from offline learning to online learning. 
We use gradient descent
with and without momentum on MinAtar \cite{minatar}, a subset of Atari games with reduced visual complexity;  MinAtar results have been shown to translate to Atari \citep{ceron2021revisiting}. We show results on Breakout in the main text, with Space Invaders results in Appendix~\ref{app:space}. Since not all the data regimes we consider allow for full-batch training, results in this section use mini-batch training; we provide full-batch training results in Appendix~\ref{app:fullbatch}. Experimental details are provided in Appendix~\ref{app:replicateres}, and training details of the pre-trained agent used in the offline RL experiments are in Appendix \ref{app:offline_rl}.

\subsection{DQN}
\begin{figure}[t]
    \centering

    \begin{subfigure}[b]{0.85\textwidth}
        \centering
        \includegraphics[width=\textwidth]{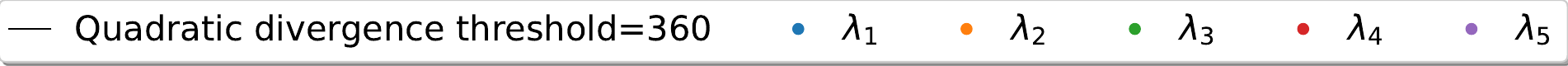}
        \label{fig:legend}
        \vspace{-0.5cm}
    \end{subfigure}
    \begin{subfigure}[b]{0.475\textwidth}
        \centering
        \includegraphics[width=\textwidth]{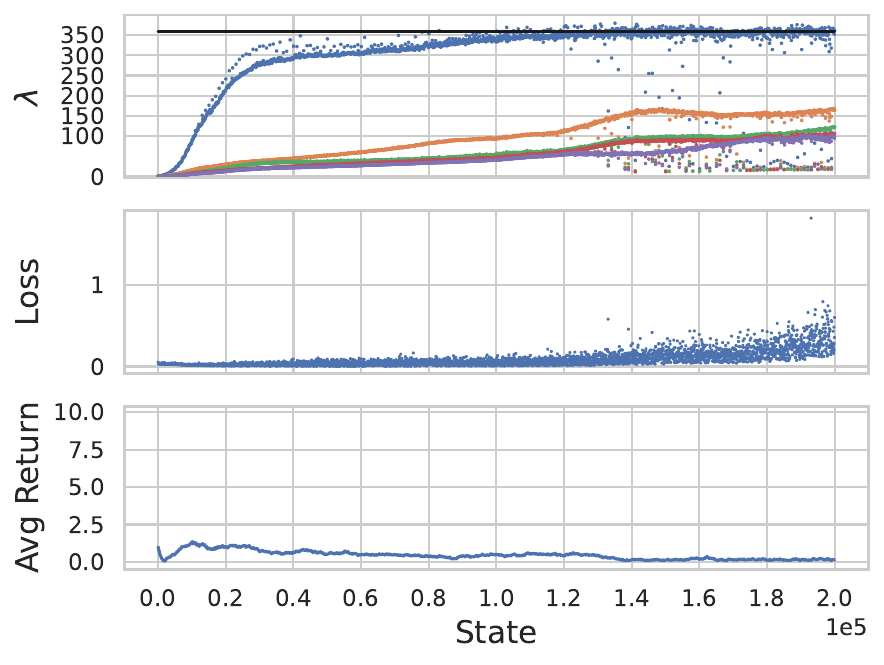}
        \caption[]%
        {\settingA.}    
        \label{fig:dqn fully offline}
    \end{subfigure}%
    \hfill
    \begin{subfigure}[b]{0.475\textwidth}  
        \centering 
        \includegraphics[width=\textwidth]{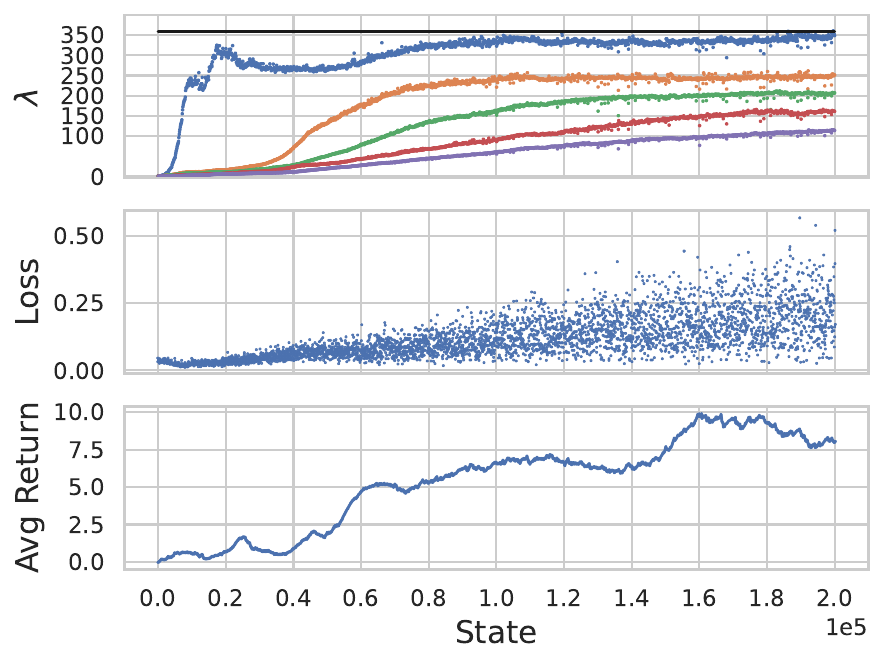}
        \caption[]%
        {\settingB.}    
        \label{fig:dqn 0.3 perturbed}
    \end{subfigure}%
    \vskip\baselineskip
    \begin{subfigure}[b]{0.85\textwidth}
        \centering
        \includegraphics[width=\textwidth]{img/legend.pdf}
        \label{fig:legend_2}
        \vspace{-0.5cm}
    \end{subfigure}
    \begin{subfigure}[b]{0.475\textwidth}   
        \centering 
        \includegraphics[width=\textwidth]{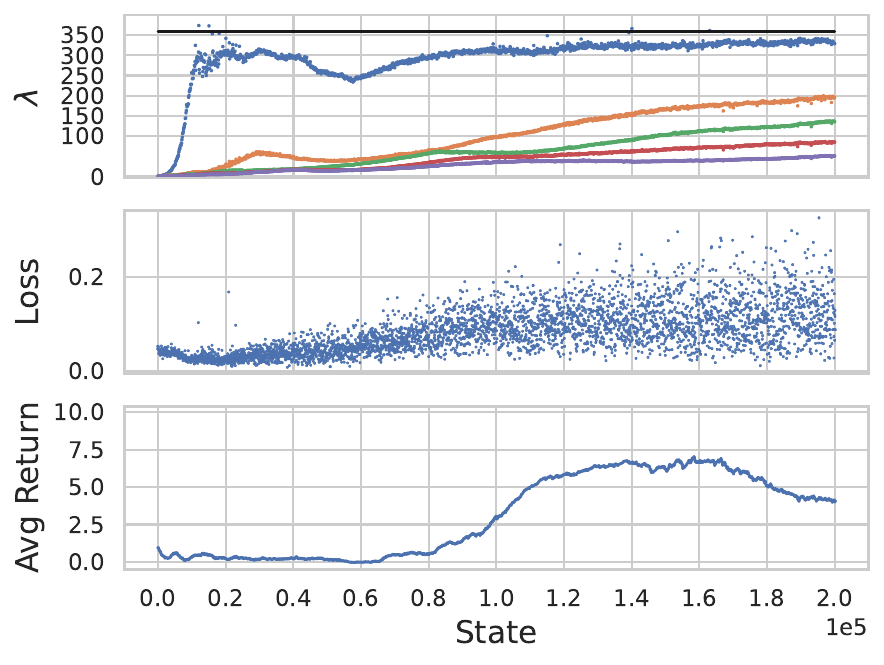}
        \caption[]%
        {\settingC.}    
        \label{fig:dqn last million}
    \end{subfigure}%
    \hfill
    \begin{subfigure}[b]{0.475\textwidth}   
        \centering 
        \includegraphics[width=\textwidth]{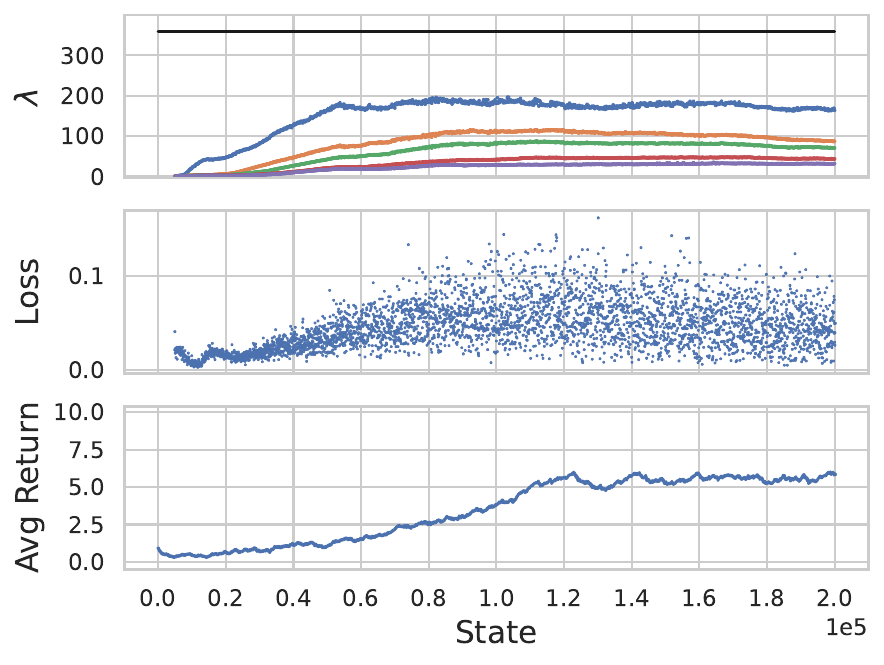}
        \caption[]%
        {\settingD.}    
        \label{fig:dqnonline}
    \end{subfigure}%
    \caption[]
    {\textbf{DQN.} The edge of stability phenomenon is observed in the offline setting (\ref{fig:dqn fully offline}, \ref{fig:dqn 0.3 perturbed}, \ref{fig:dqn last million}): the leading eigenvalue $\lambda_1$  rises to the quadratic divergence threshold induced by the underlying optimisation algorithm, after which it fluctuates around the threshold; when $\lambda_1$ reaches the quadratic divergence threshold, we observe more loss instability, but this does not always get reflected in the agent's reward. In online training, $\lambda_1$ does not reach the quadratic divergence threshold and we do not observe the edge of stability phenomenon, though $\lambda_1$ plateaus early in training.} 
    \label{fig:dqnalldatasets}
\end{figure}

\textbf{\settingA}. We use a pre-trained agent's greedy policy to generate a replay buffer of $10^6$ transitions and use this to train a new DQN agent. This setup is closest to that of supervised learning, and isolates the effect of the RL losses and bootstrapping from RL specific effects on the data distribution. Figure \ref{fig:dqn fully offline} shows a clear edge of stability effect: the leading eigenvalue $\lambda_1$ grows until reaching the quadratic divergence threshold, after which fluctuates around it and the loss function shows increasing instabilities; this is consistent with results using the mean squared error in supervised learning.  Consistent with existing results \cite{offline_rl_no_exploration}, the performance of the agent is poor, likely due distributional shift. 

\textbf{\settingB}. To address the challenges with distributional shift and increase the diversity of agent experience, instead using greedy actions  taken by the pre-trained agent to generate the replay buffer, we use an $\epsilon$-greedy policy with 0.7 probability of taking greedy action from the pre-trained agent and 0.3 probability of a random action. This is akin to the concurrent setting in \citet{offline_rl_no_exploration} . Results in Figure \ref{fig:dqn 0.3 perturbed} show that this intervention vastly improves the reward obtained by the agent compared to the previous setting explored, but optimisation dynamics retain the edge of stability behaviour.

\textbf{\settingC}. We train an agent using the last $10^6$ transitions obtained from the pre-trained agent's online phase; this is known as the {final buffer} setting \cite{offline_rl_no_exploration,optimistic_offline_rl}. This brings us closer to online RL: we still use a fixed replay buffer to train the agent, but that dataset contains transitions from a changing distribution. This allows us to isolate the effect of the replay buffer being obtained from a series of changing policies from the interactions of the agent's behaviour affecting its own replay buffer, as we see in online learning. Results in Figure \ref{fig:dqn last million} show that here too, we observe the edge of stability phenomenon. 

\textbf{\settingD}. In online RL, the replay buffer is obtained from the agent's own experience, leading to the related challenges mentioned in Section~\ref{sec:challenges_rl}. Figure \ref{fig:dqnonline} shows that the leading eigenvalues of the Hessian grow early in training but plateau below the quadratic divergence threshold; despite $\lambda_1$ not reaching the quadratic divergence threshold, once it plateaus we observe increased instability in the loss function.

We show full-batch results across the above offline RL cases above in Figures~\ref{fig:dqn_000_full_batch_zoom},~\ref{fig:dqn_030_full_batch_zoom},~\ref{fig:dqn_last_mil_full_batch_zoom} in the Appendix, which consistently show that as $\lambda_1$ fluctuates around the quadratic divergence threshold the loss function exhibits increased instabilities. We note that while changing the target network can have a short term effect on $\lambda_1$, it does not drastically affect it's long term trajectory and the edge of stability phenomenon. 

\subsection{C51}
\begin{figure}[t]
    \centering

    \begin{subfigure}[b]{0.475\textwidth}   
        \centering 
        \includegraphics[width=\textwidth]{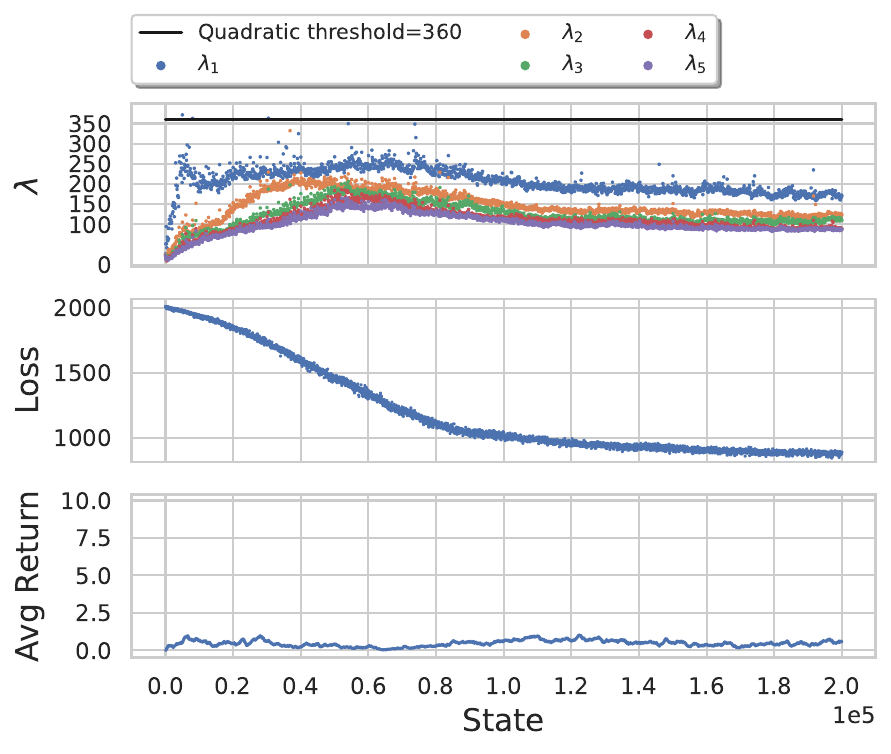}
        \caption[]%
        {\settingA.}    
        \label{fig:c51_offline_main}
    \end{subfigure}
    \hfill
    \begin{subfigure}[b]{0.475\textwidth}   
        \centering 
        \includegraphics[width=\textwidth]{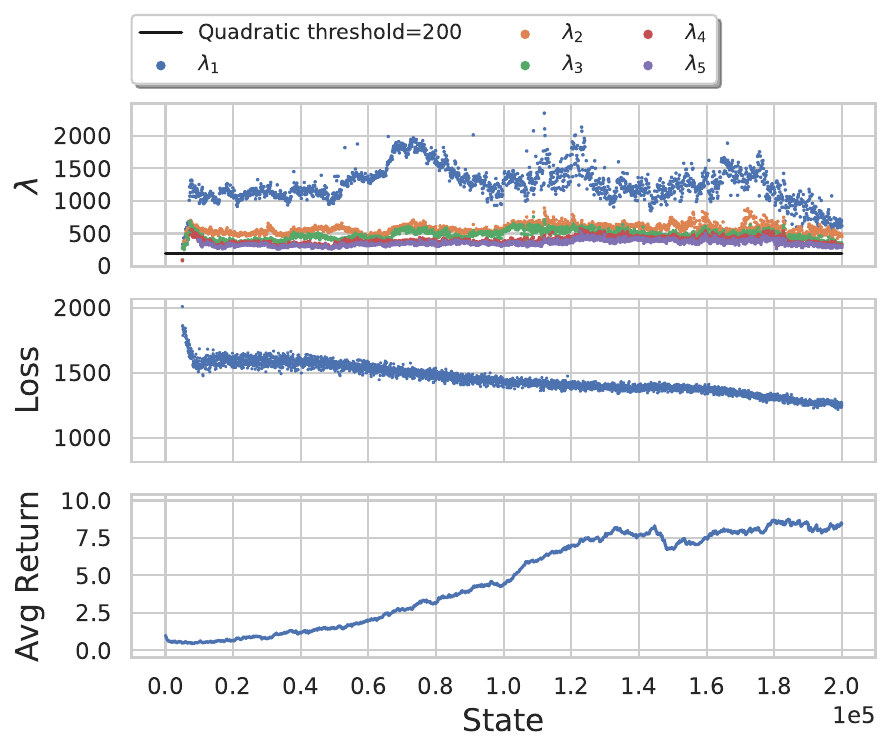}
        \caption[]%
        {\settingD.}    
        \label{fig:c51_online_main}
    \end{subfigure}
    \caption[]
    {\textbf{C51}. When using C51 in an offline setting, we notice that the leading eigenvalue $\lambda_1$ does not always reach the edge of stability threshold. In online learning, we observe that $\lambda_1$ greatly exceeds the quadratic divergence threshold. Like in supervised learning, however, using a cross entropy loss leads to a decrease of $\lambda_1$ later in training.} 
    \label{fig:c51_main}
\end{figure}

When investigating the edge of stability effect using C51 across all the above data regimes, we observe that C51 does not clearly exhibit the edge of stability behaviour; we show selected results in Figure~\ref{fig:c51_main} and additional results in Figures~\ref{fig:c51alldatasets} and \ref{fig:c51alldatasets_sgd} in the Appendix. However, similar to the observations in supervised learning with cross-entropy loss~\citep{edgeofstability}, we notice that $\lambda_1$ grows early in training, after which it consistently decreases. In offline learning (Figure~\ref{fig:c51_offline_main}), the leading eigenvalue $\lambda_1$, usually stays under the quadratic divergence threshold, but this is not the case in online learning (Figure~\ref{fig:c51_online_main}), where $\lambda_1$ is consistently significantly above the quadratic divergence threshold. This observation might explain why we observed increased challenges with training C51 in this setting compared to DQN (results in Figure~\ref{fig:c51_online_main} use stochastic gradient descent without momentum, as using momentum led to very poor results, see Figure \ref{fig:c51online} in the Appendix).

\section{Discussion}
We examined the edge of stability effect in DQN and C51, two off-policy deep RL algorithms on simple environments. Our findings suggest that  \textit{the edge of stability phenomenon can be induced by neural network optimisation in deep RL, but whether this occurs depends on underlying algorithm}. We observed that DQN exhibits the edge of stability behaviour in offline RL, with a diminished effect in online RL. In contrast, we did not observe a consistent edge of stability effect when using C51, but nonetheless did observe a connection between large leading Hessian eigenvalues and challenges in training C51 agents.

\textbf{Caveats and future work}. 
Our results were obtained on the MinAtar environment; we hope that future studies will expand our results to a wider range of environments. Following~\citet{edgeofstability}, we investigate the edge of stability phenomenon in RL when using gradient descent with momentum; we believe future work can expand our exploration to adaptive optimisers commonly used in RL, such as Adam \citep{kingma2014adam}, as has recently been done in supervised learning \citep{cohen2022adaptive}. We further hope future research can connect the leading eigenvalue of the Hessian to the agent's performance, not only the loss, as has been done in supervised learning with generalisation \citep{hochreiter1997flat,keskarlarge,jastrzkebskirelation,lewkowycz2020large}.

\clearpage

\vskip 0.2in
\bibliography{sample}

\newpage

\appendix
\section{Additional experimental results}\label{app:additional_exp}

\subsection{SGD with momentum results for C51 on Breakout}
\begin{figure}[H]
    \centering
    \begin{subfigure}[b]{0.475\textwidth}
        \centering
        \includegraphics[width=\textwidth]{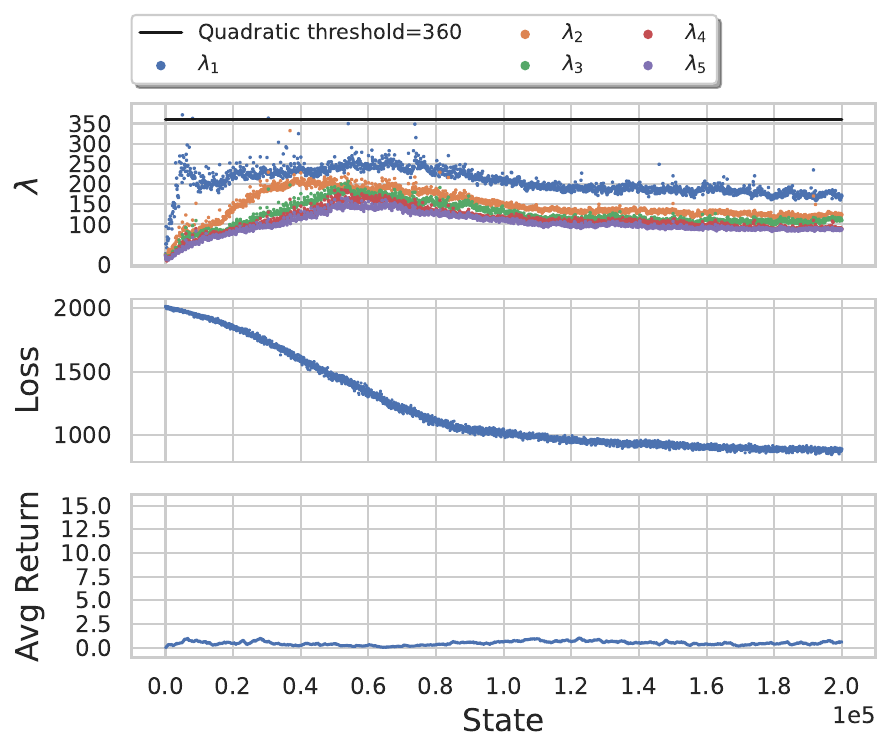}
        \caption[]%
        {\settingA.}  
        \label{fig:c51 fully offline}
    \end{subfigure}
    \hfill
    \begin{subfigure}[b]{0.475\textwidth}  
        \centering 
        \includegraphics[width=\textwidth]{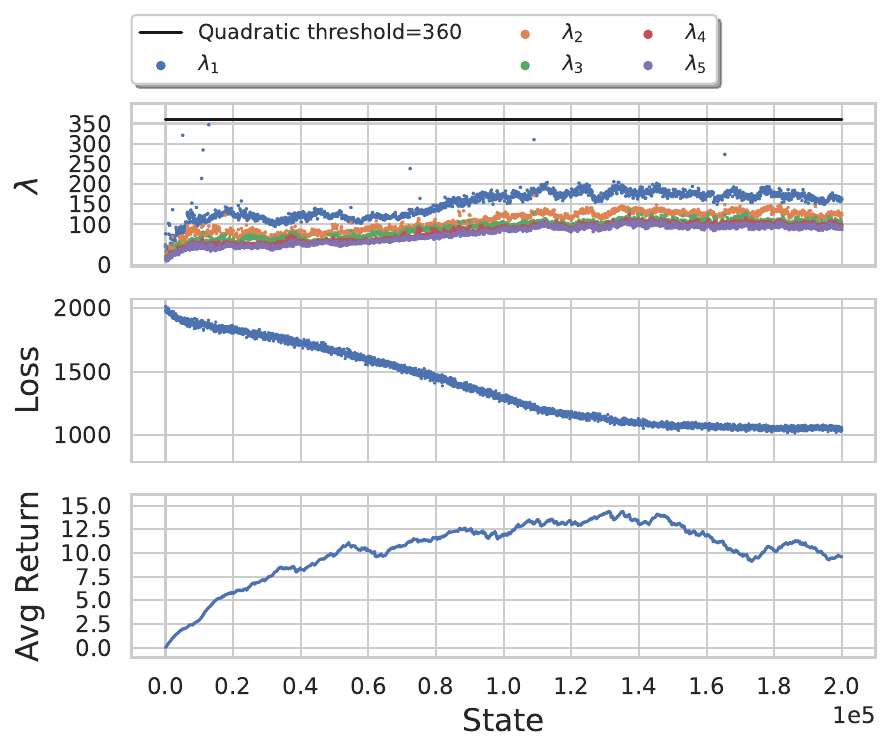}
        \caption[]%
        {\settingB.}       
        \label{fig:c51 0.3 perturbed}
    \end{subfigure}
    \vskip\baselineskip
    \begin{subfigure}[b]{0.475\textwidth}   
        \centering 
        \includegraphics[width=\textwidth]{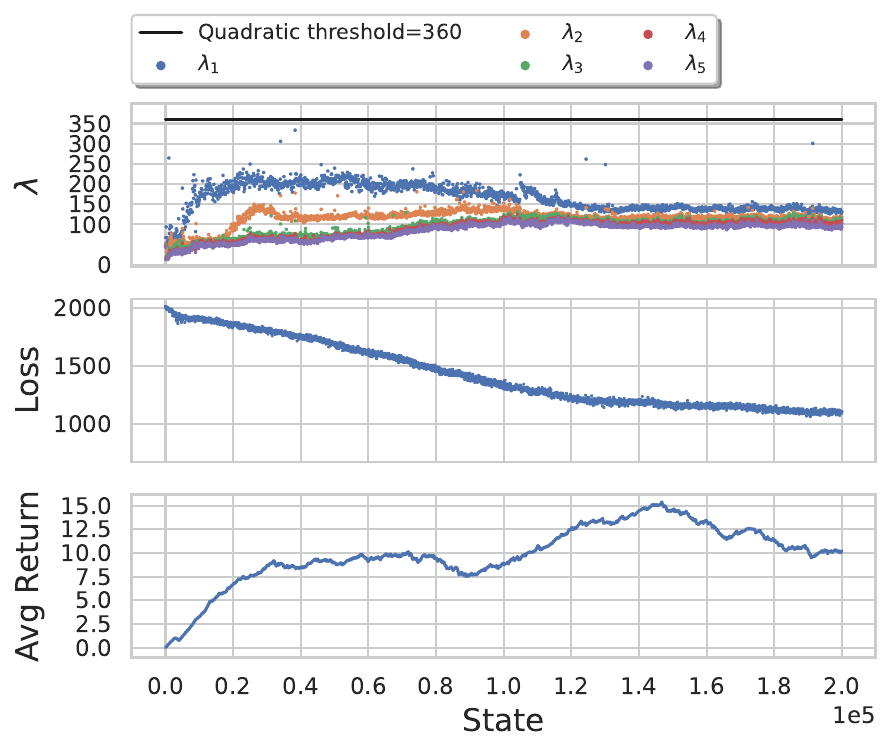}
        \caption[]%
        {\settingC.}    
        \label{fig:c51 last million}
    \end{subfigure}
    \hfill
    \begin{subfigure}[b]{0.475\textwidth}   
        \centering 
        \includegraphics[width=\textwidth]{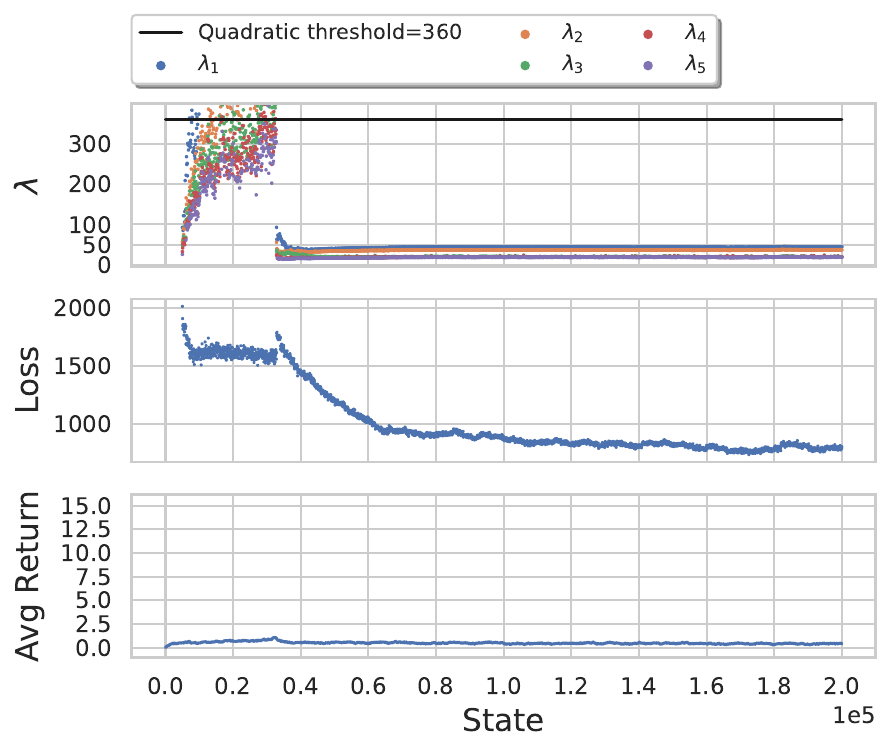}
        \caption[]%
        {\settingD.}   
        \label{fig:c51online}
    \end{subfigure}
    \caption[]
    {\textbf{C51}. When using C51 in an offline setting, we notice that the leading eigenvalue $\lambda_1$ does not always reach the edge of stability threshold. In online learning, we observe that $\lambda_1$ greatly exceeds the quadratic divergence threshold at the beginning of training after which it loses orders of magnitude and plateaus at a level below the quadratic threshold. Like in supervised learning, however, using a cross entropy loss leads to a decrease of $\lambda_1$ later in training.}  
    \label{fig:c51alldatasets}
\end{figure}

In Figure~\ref{fig:c51alldatasets} we present C51 results on Breakout which do not clearly show an edge of stability effect. In the online regime, the eigenvalues consistently rise orders of magnitude above the quadratic threshold but reach low levels and plateau later in training.

\subsection{Experiments on Breakout using SGD with momentum and full-batch}\label{app:fullbatch}

\begin{figure}[H]
    \centering
    \begin{subfigure}[b]{0.475\textwidth}
        \centering
        \includegraphics[width=\textwidth]{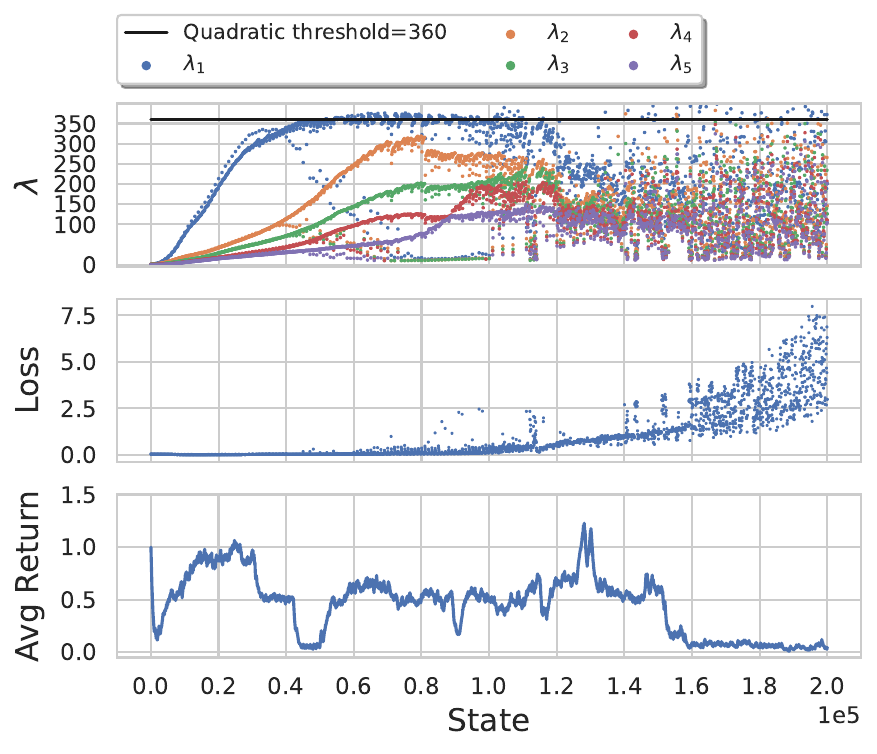}
        \caption[]%
        {{DQN.}}    
        \label{fig:dqnfullbatchoffline}
    \end{subfigure}
    \hfill
    \begin{subfigure}[b]{0.475\textwidth}  
        \centering 
        \includegraphics[width=\textwidth]{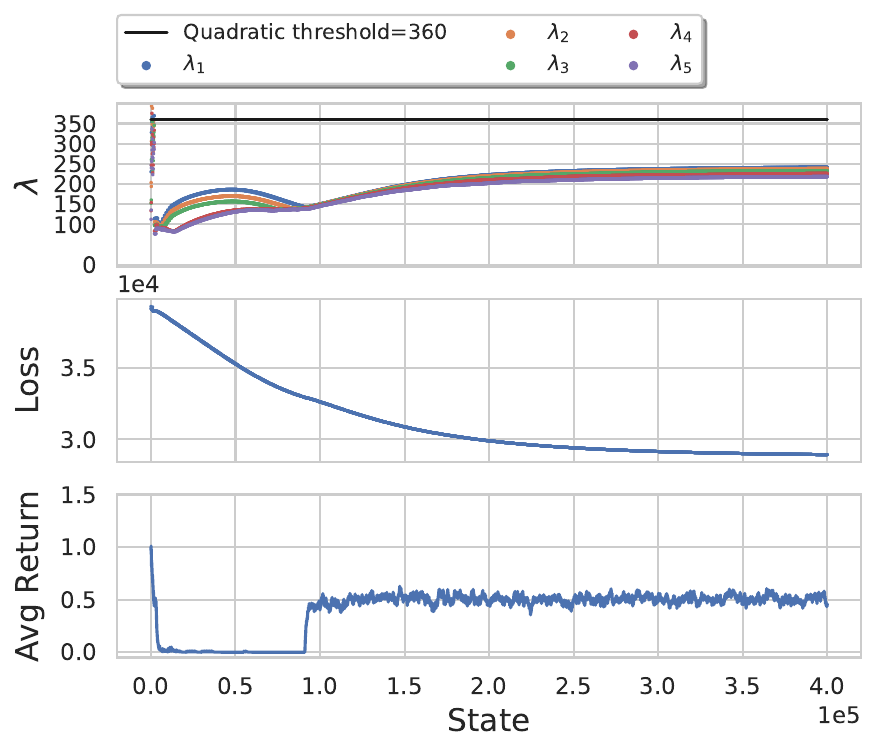}
        \caption[]%
        {{C51.}}    
        \label{fig:c51fullbatchoffline}
    \end{subfigure}
    
    \caption[]
    {Full batch results in the setting \settingA. \textbf{DQN} shows a strong edge of stability effect followed by instabilities in the eigenvalues. \textbf{C51} presents a rise and fall trend initially but over time the eigenvalues plateau.} 
    \label{fig:fullbatchoffline}
\end{figure}

\begin{figure}[H]
    \centering
    \includegraphics[width=0.5\textwidth]{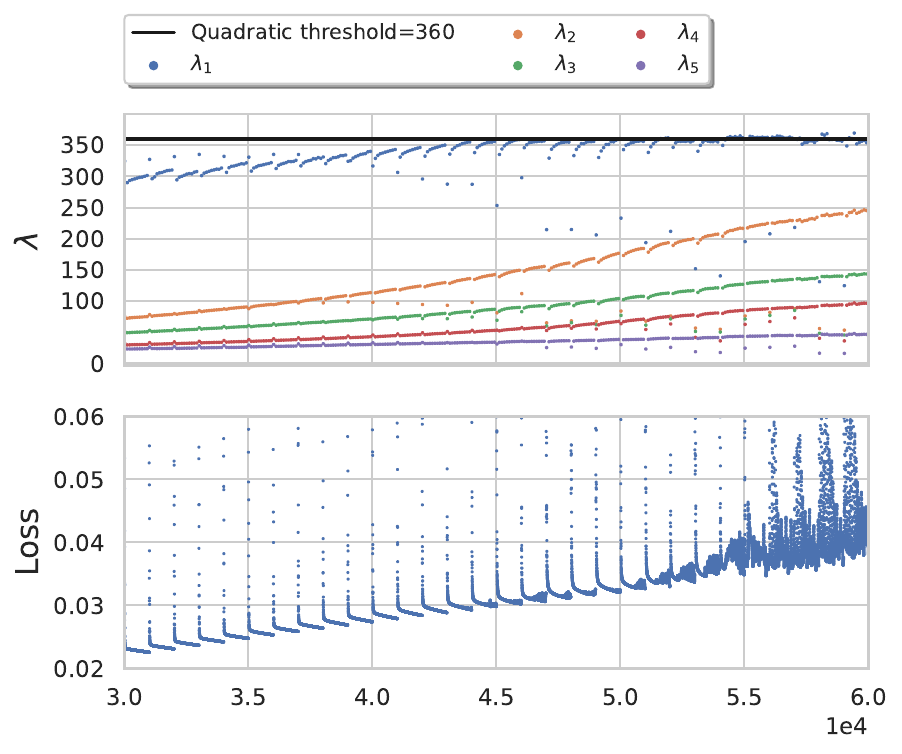}
    \caption{Zoom in full-batch results in the setting \settingA. \textbf{DQN} shows a strong edge of stability effect. The behaviour is consistent with supervised learning where the loss and leading eigenvalue have the same relationship across changes to the target network.}
    \label{fig:dqn_000_full_batch_zoom}
\end{figure}

In Figure~\ref{fig:fullbatchoffline} we present full-batch experiments for DQN and C51 in the setting \settingA, with a zoom in on the first time the quadratic threshold is achieved in Figure~\ref{fig:dqn_000_full_batch_zoom}. DQN shows a clear edge of stability effect which is broken later during training where we see increased instabilities. C51 does not show an edge of stability effect with the eigenvalues plateauing over time.

\begin{figure}[H]
    \centering
    \begin{subfigure}[b]{0.475\textwidth}
        \centering
        \includegraphics[width=\textwidth]{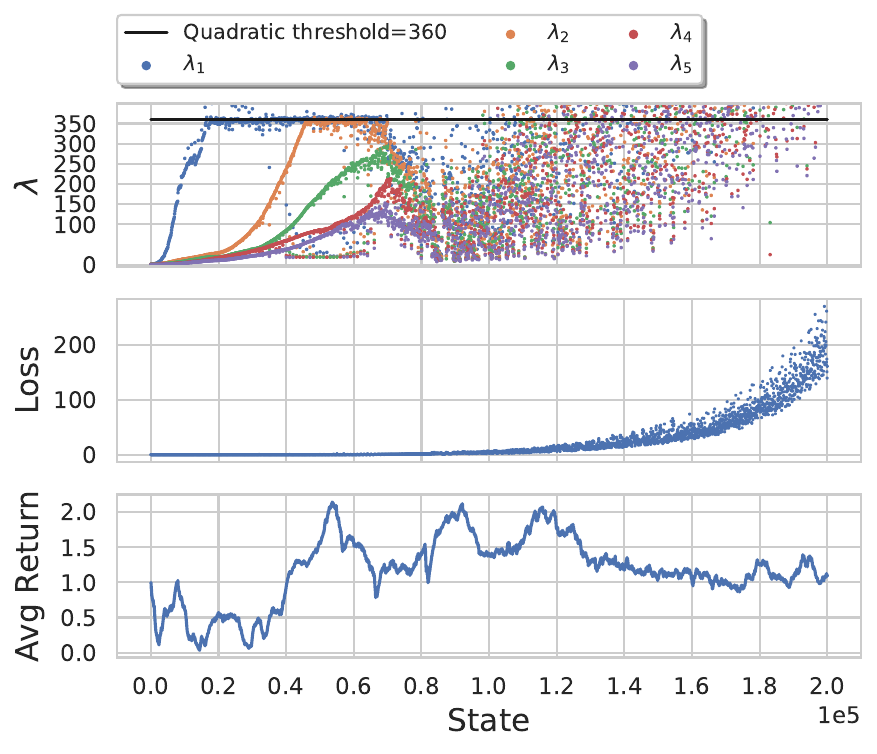}
        \caption[]%
        {{DQN.}}    
        \label{fig:dqnfullbatchoffline030}
    \end{subfigure}
    \hfill
    \begin{subfigure}[b]{0.475\textwidth}  
        \centering 
        \includegraphics[width=\textwidth]{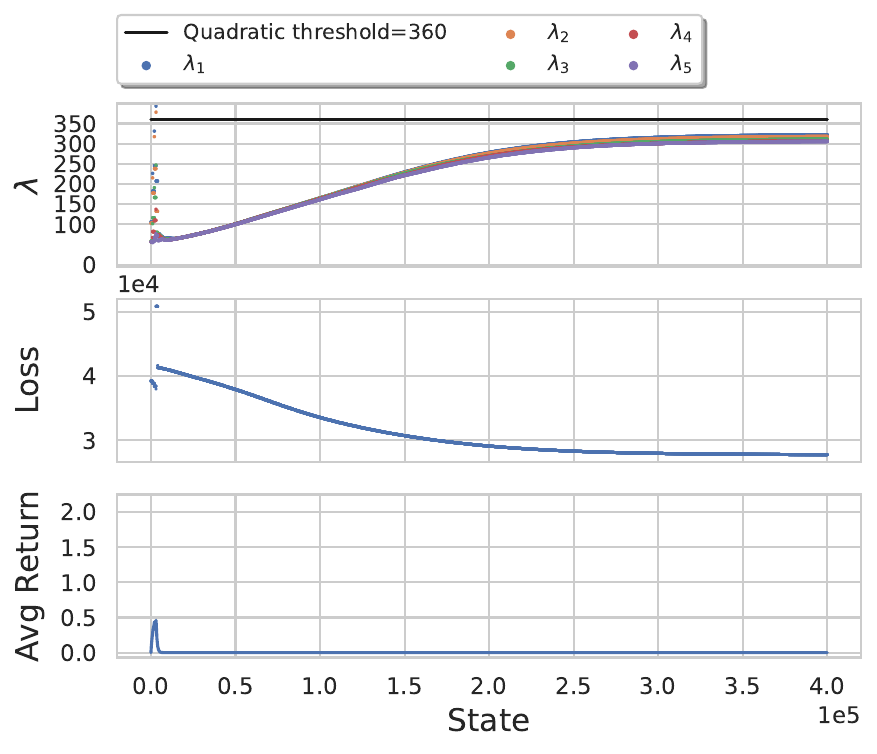}
        \caption[]%
        {{C51.}}    
        \label{fig:c51fullbatchoffline030}
    \end{subfigure}
    
    \caption[]
    {Full batch results in the setting \settingB. \textbf{DQN} shows a strong edge of stability effect followed by instabilities in the eigenvalues. In the case of \textbf{C51} the eigenvalues plateau over time.}
    \label{fig:fullbatchoffline030}
\end{figure}

\begin{figure}[H]
    \centering
    \includegraphics[width=0.5\textwidth]{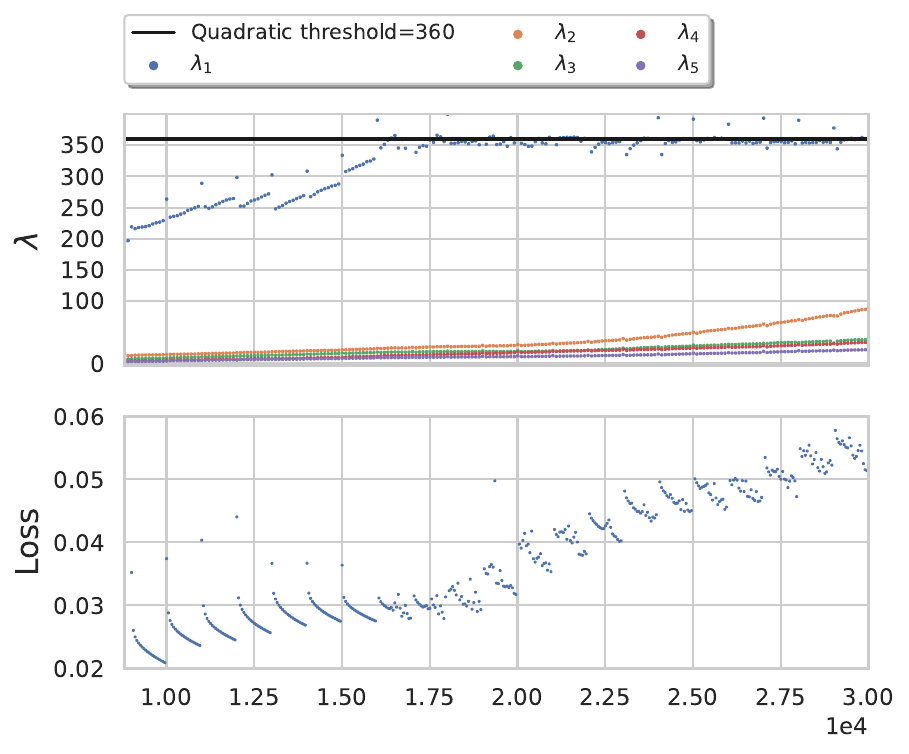}
    \caption{Zoom in full-batch results in the setting \settingB. \textbf{DQN} shows a strong edge of stability effect. The behaviour is consistent with supervised learning where the loss and leading eigenvalue have the same relationship across changes to the target network.}
    \label{fig:dqn_030_full_batch_zoom}
\end{figure}

In Figure~\ref{fig:fullbatchoffline030} we present full-batch experiments for DQN and C51 in the setting \settingB, with a zoom in on the first time the quadratic threshold is achieved in Figure~\ref{fig:dqn_030_full_batch_zoom}. Similar to the previous setting, DQN shows a clear edge of stability effect which is broken later during training where we see increased instabilities. C51 does not show an edge of stability effect with the eigenvalues plateauing over time.

\begin{figure}[H]
    \centering
    \begin{subfigure}[b]{0.475\textwidth}
        \centering
        \includegraphics[width=\textwidth]{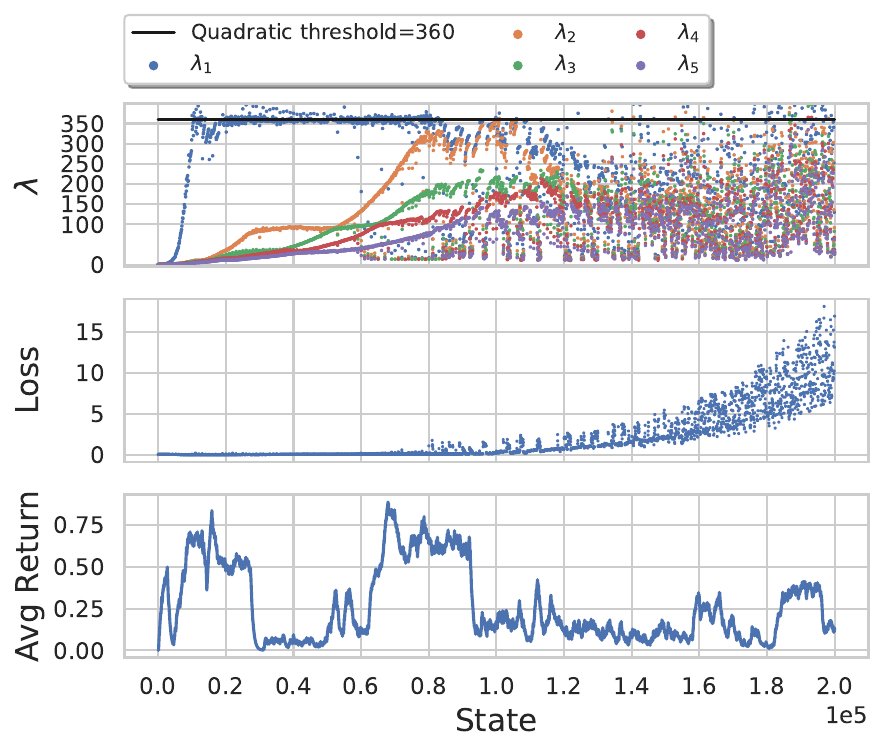}
        \caption[]%
        {{DQN.}}    
        \label{fig:dqnfullbatchlastmillion}
    \end{subfigure}
    \hfill
    \begin{subfigure}[b]{0.475\textwidth}  
        \centering 
        \includegraphics[width=\textwidth]{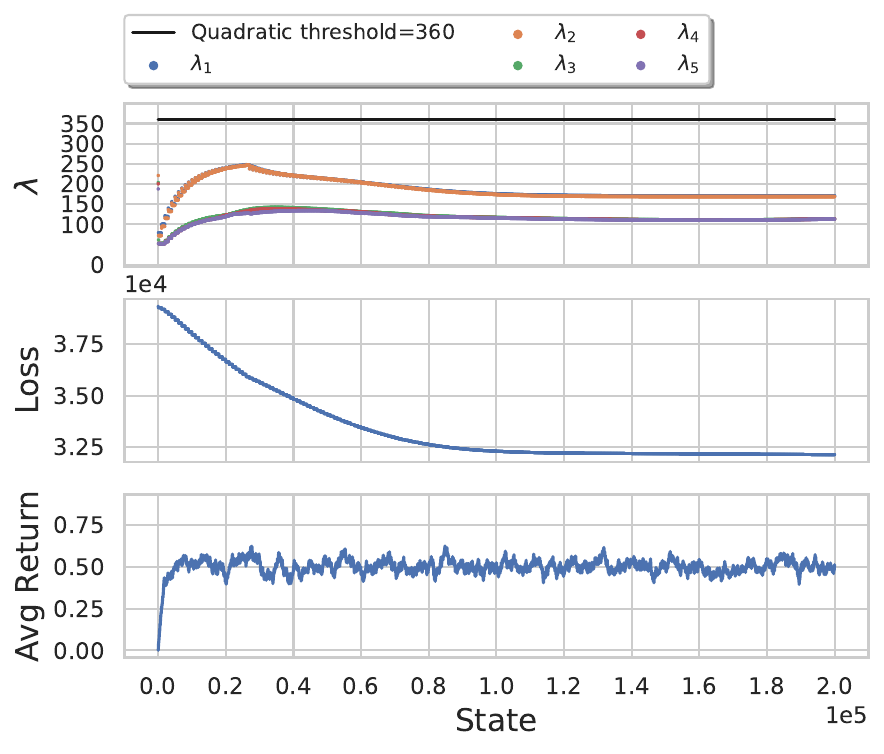}
        \caption[]%
        {{C51.}}    
        \label{fig:c51fullbatchlastmillion}
    \end{subfigure}
    
    \caption[]
    {Full batch results in the setting \settingC. \textbf{DQN} shows a strong edge of stability effect. The behaviour is consistent with supervised learning where the loss and leading eigenvalue have the same relationship across changes to the target network.}
    \label{fig:fullbatchlast_million}
\end{figure}

\begin{figure}[H]
    \centering
    \includegraphics[width=0.5\textwidth]{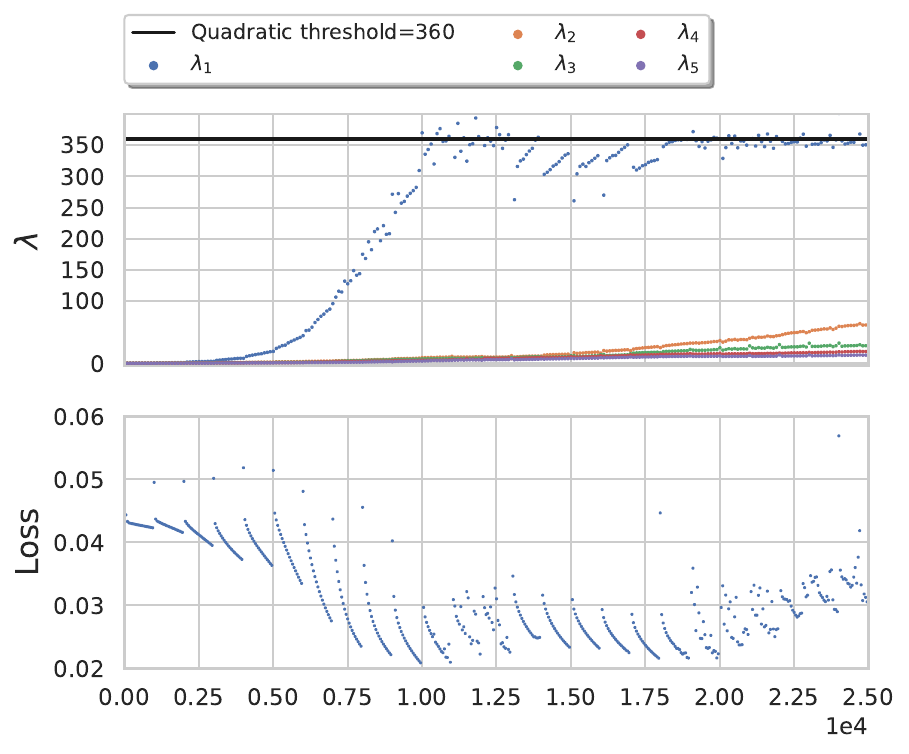}
    \caption{Zoom in full-batch results in the setting \settingC. \textbf{DQN} shows a strong edge of stability effect. The behaviour is consistent with supervised learning where the loss and leading eigenvalue have the same relationship across changes to the target network.}
    \label{fig:dqn_last_mil_full_batch_zoom}
\end{figure}

In Figure~\ref{fig:fullbatchlast_million} we present full-batch experiments for DQN and C51 in the setting \settingC, with a zoom in on the first time the quadratic threshold is achieved in Figure~\ref{fig:dqn_last_mil_full_batch_zoom}. Similar to the previous two settings, DQN shows a clear edge of stability effect which is broken later during training where we see increased instabilities. C51 does not show an edge of stability effect with the eigenvalues plateauing over time.

\subsection{SGD results for DQN and C51 on Breakout}
\begin{figure}[H]
    \centering
    \begin{subfigure}[b]{0.475\textwidth}
        \centering
        \includegraphics[width=\textwidth]{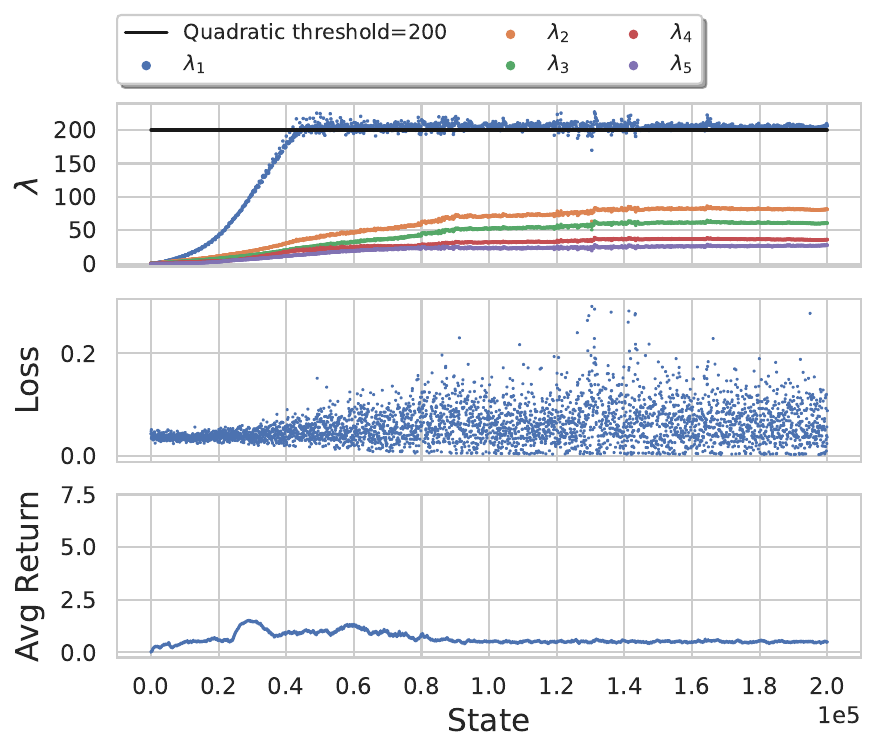}
        \caption[]%
        {\settingA.}    
        \label{fig:dqn fully offline_sgd}
    \end{subfigure}
    \hfill
    \begin{subfigure}[b]{0.475\textwidth}  
        \centering 
        \includegraphics[width=\textwidth]{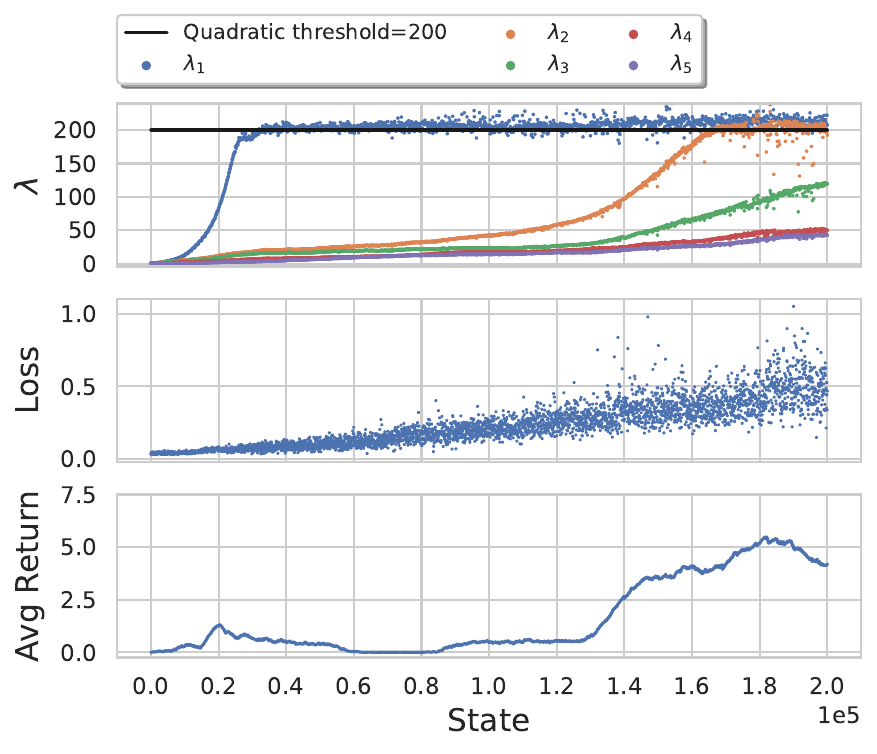}
        \caption[]%
        {\settingB.}    
        \label{fig:dqn 0.3 perturbed_sgd}
    \end{subfigure}
    \vskip\baselineskip
    \begin{subfigure}[b]{0.475\textwidth}   
        \centering 
        \includegraphics[width=\textwidth]{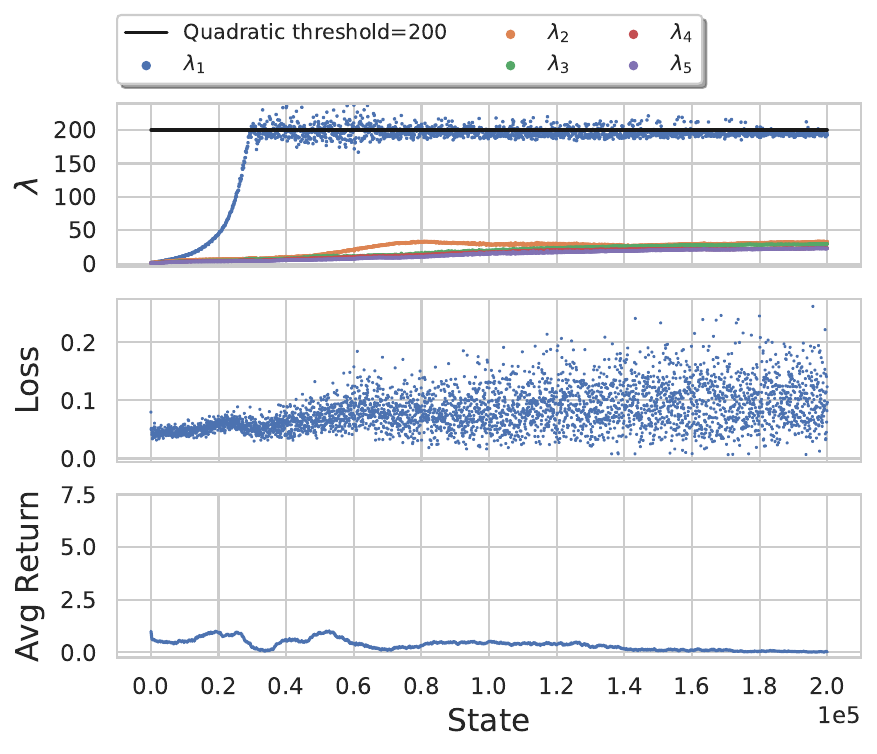}
        \caption[]%
        {\settingC.}    
        \label{fig:dqn last million_sgd}
    \end{subfigure}
    \hfill
    \begin{subfigure}[b]{0.475\textwidth}   
        \centering 
        \includegraphics[width=\textwidth]{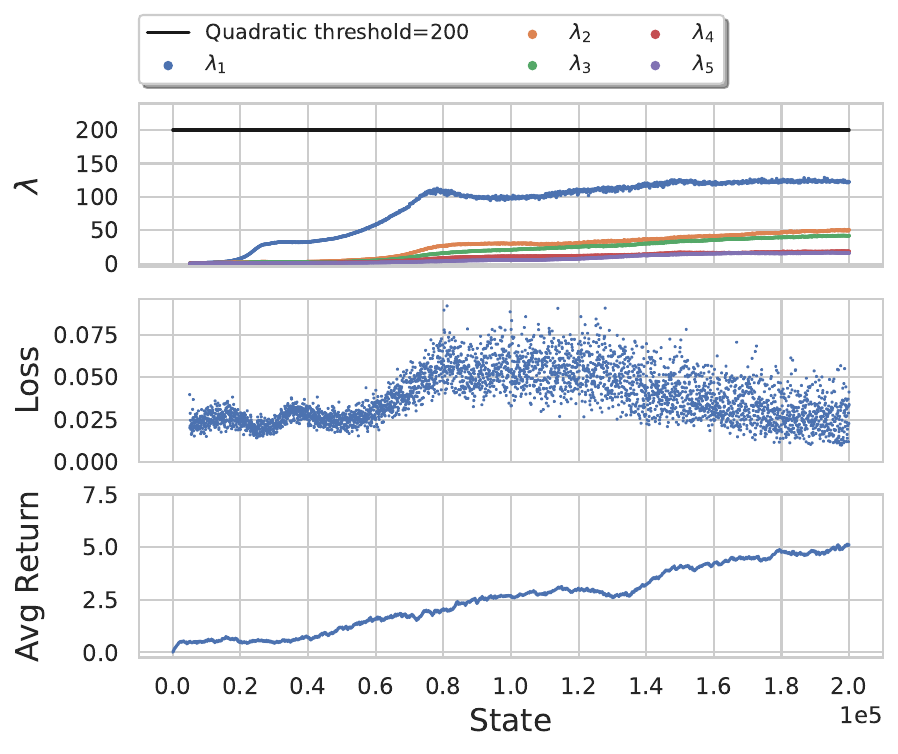}
        \caption[]%
        {\settingD.}    
        \label{fig:dqnonline_sgd}
    \end{subfigure}
    \caption[]
    {\textbf{DQN.} The edge of stability phenomenon is observed in the offline setting (\ref{fig:dqn fully offline_sgd}, \ref{fig:dqn 0.3 perturbed_sgd} and \ref{fig:dqn last million_sgd}): the leading eigenvalue $\lambda_1$  rises to the quadratic threshold, after which it hovers around the quadratic threshold, albeit with some noise; when $\lambda_1$ reaches the quadratic divergence threshold, we observe more loss instability, but this does not always get reflected in the agent's reward. In online training, $\lambda_1$ does not reach the quadratic divergence threshold. } 
    \label{fig:dqnalldatasets_sgd}
\end{figure}

In Figure~\ref{fig:dqnalldatasets_sgd} we present DQN results on Breakout with SGD without momentum which do not clearly show an edge of stability effect. In the online regime, the eigenvalues consistently fail to rise to the quadratic threshold but reach low levels and plateau later in training. There is a clear edge of stability effect offline.

\begin{figure}[H]
    \centering
    \begin{subfigure}[b]{0.475\textwidth}
        \centering
        \includegraphics[width=\textwidth]{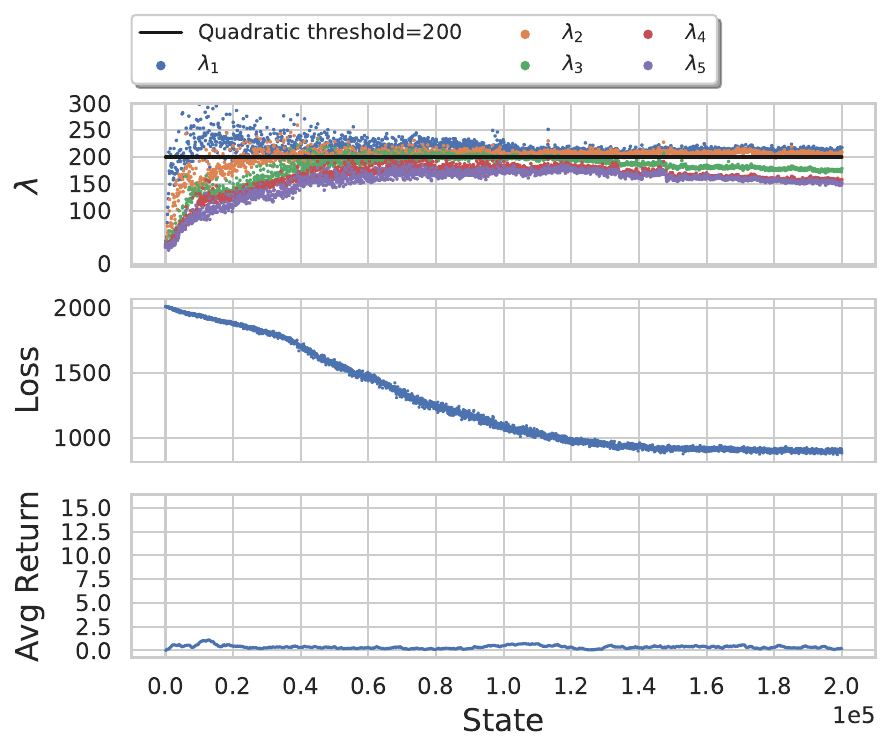}
        \caption[]%
        {\settingA.}    
        \label{fig:c51 fully offline_sgd}
    \end{subfigure}
    \hfill
    \begin{subfigure}[b]{0.475\textwidth}  
        \centering 
        \includegraphics[width=\textwidth]{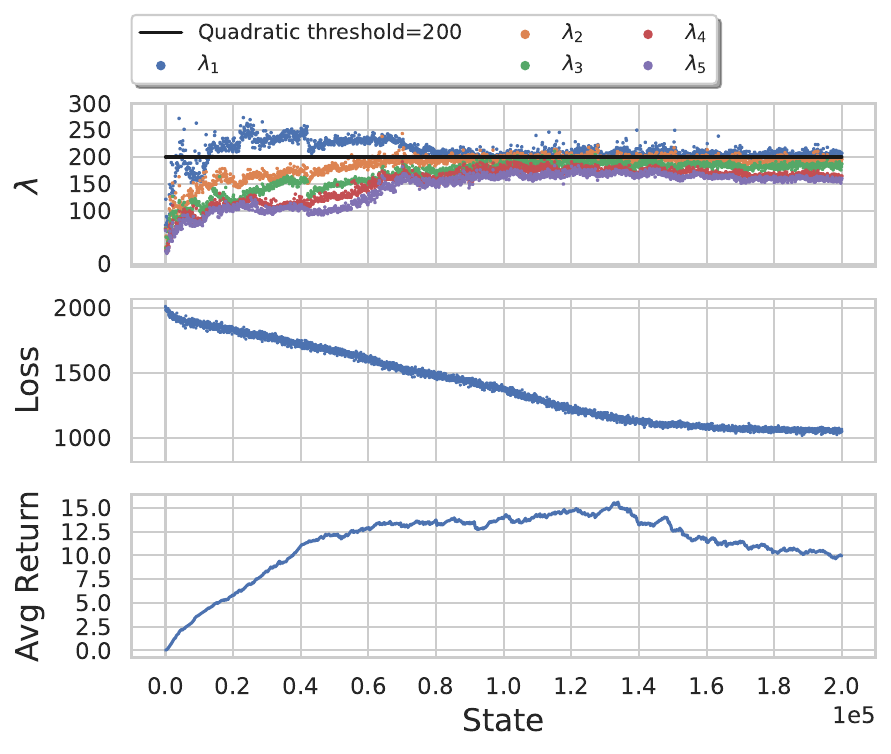}
        \caption[]%
        {\settingB.}    
        \label{fig:c51 0.3 perturbed_sgd}
    \end{subfigure}
    \vskip\baselineskip
    \begin{subfigure}[b]{0.475\textwidth}   
        \centering 
        \includegraphics[width=\textwidth]{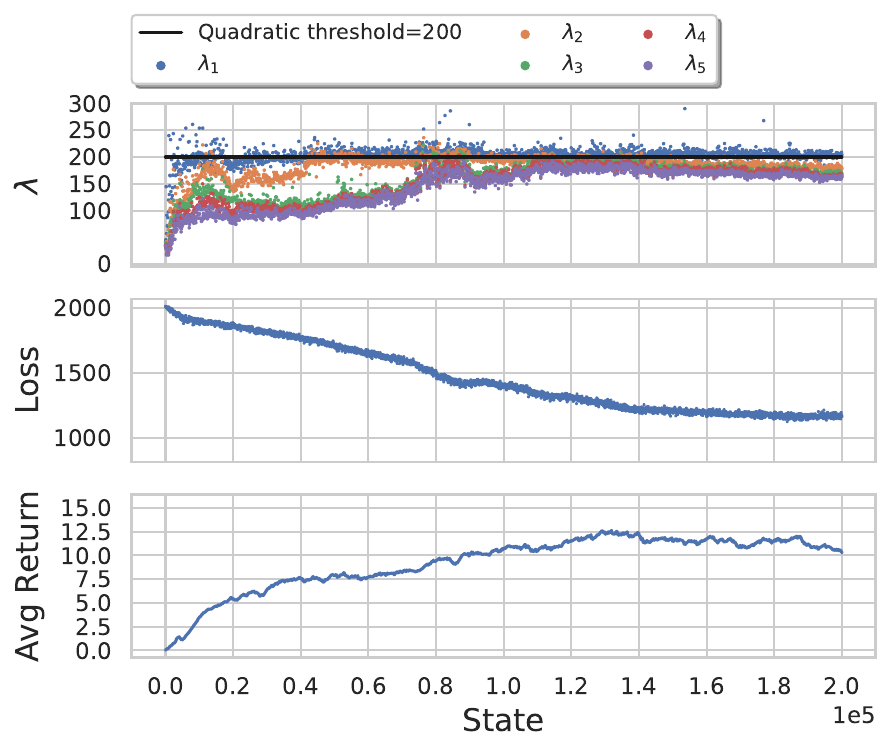}
        \caption[]%
        {\settingC.}    
        \label{fig:c51 last million_sgd}
    \end{subfigure}
    \hfill
    \begin{subfigure}[b]{0.475\textwidth}   
        \centering 
        \includegraphics[width=\textwidth]{img/c51/online/c51_online_sgd.pdf}
        \caption[]%
        {\settingD.}    
        \label{fig:c51online_sgd}
    \end{subfigure}
    \caption[]
    {\textbf{C51}. When using C51 in an offline setting with SGD, we notice that the leading eigenvalue $\lambda_1$ now reaches the edge of stability threshold. In online learning, we observe that $\lambda_1$ greatly exceeds the quadratic divergence threshold. Like in supervised learning, however, using a cross entropy loss leads to a decrease of $\lambda_1$ later in training.}  
    \label{fig:c51alldatasets_sgd}
\end{figure}

In Figure~\ref{fig:c51alldatasets_sgd} we present C51 results on Breakout with SGD without momentum which show an edge of stability effect. In the offline regimes, the eigenvalues rise sligtly past the quadratic threshold and then decrease to hover around it. In the online regime, the eigenvalues consistently rise orders of magnitude above the quadratic threshold.

\newpage

\subsection{Results for DQN in the Space Invaders environment with and without momentum}\label{app:space}

\begin{figure}[H]
    \centering
    \begin{subfigure}[b]{0.475\textwidth}
        \centering
        \includegraphics[width=\textwidth]{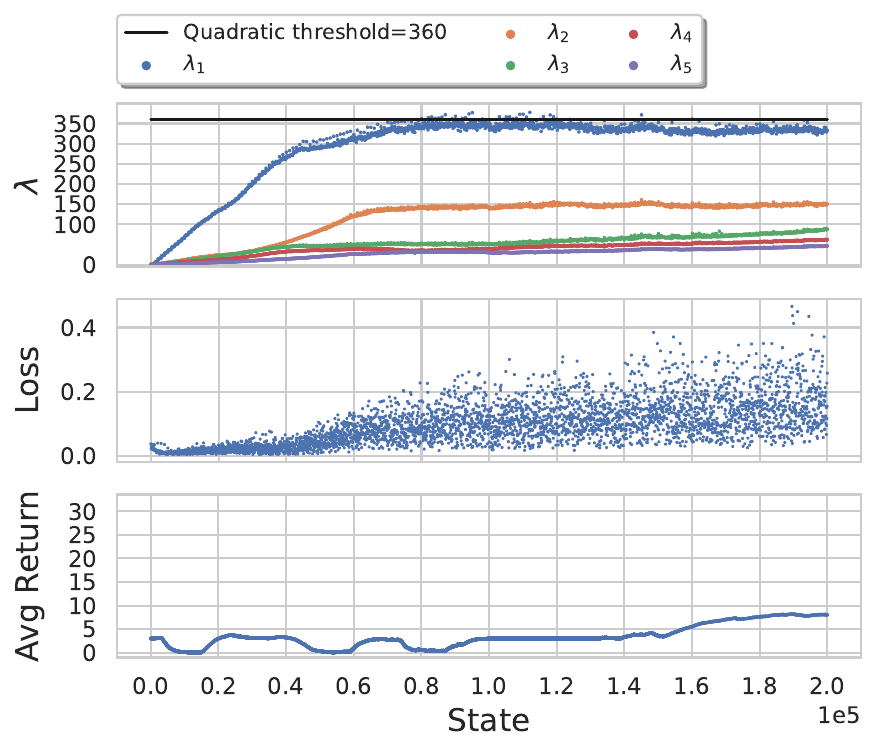}
        \caption[]%
        {\settingA.}    
        \label{fig:dqn fully offline_space}
    \end{subfigure}
    \hfill
    \begin{subfigure}[b]{0.475\textwidth}  
        \centering 
        \includegraphics[width=\textwidth]{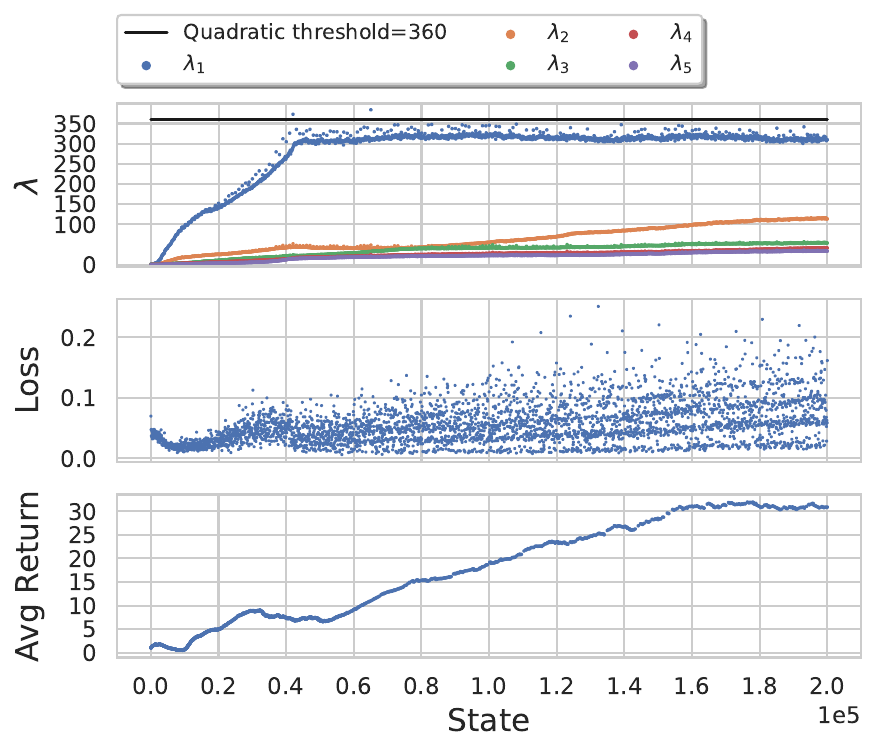}
        \caption[]%
        {\settingB.}    
        \label{fig:dqn 0.3 perturbed_space}
    \end{subfigure}
    \vskip\baselineskip
    \begin{subfigure}[b]{0.475\textwidth}   
        \centering 
        \includegraphics[width=\textwidth]{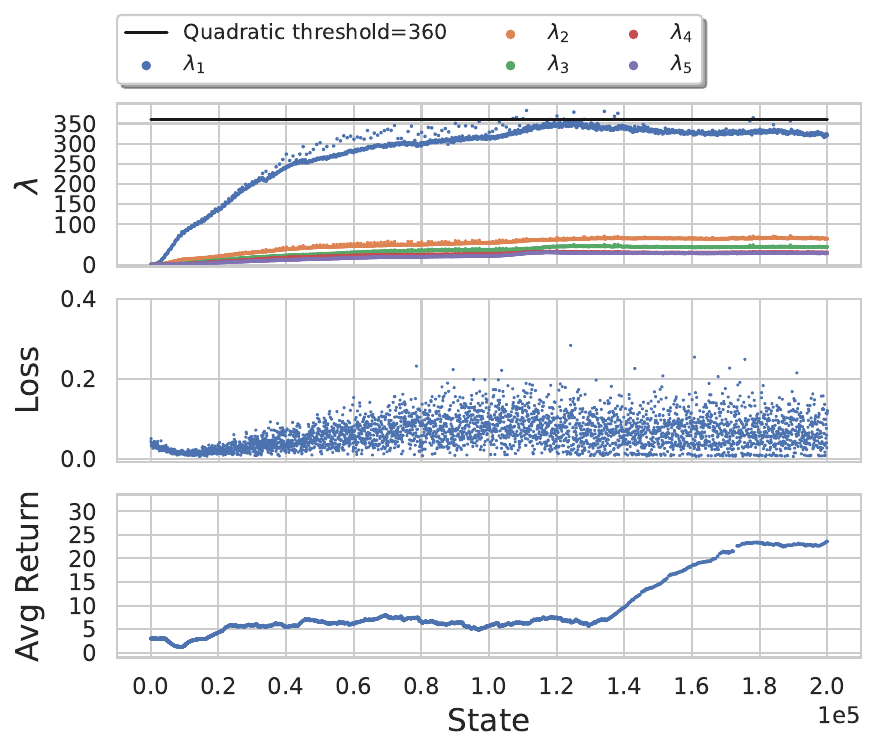}
        \caption[]%
        {\settingC.}    
        \label{fig:dqn last million_space}
    \end{subfigure}
    \hfill
    \begin{subfigure}[b]{0.475\textwidth}   
        \centering 
        \includegraphics[width=\textwidth]{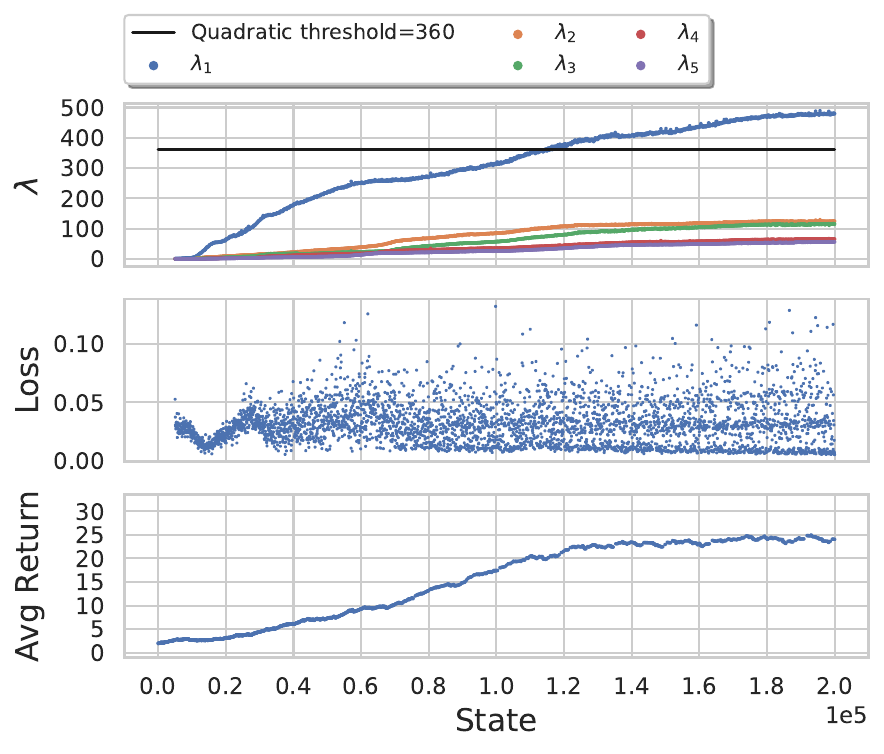}
        \caption[]%
        {\settingD.}    
        \label{fig:dqnonline_space}
    \end{subfigure}
    \caption[]
    {\textbf{DQN.} The edge of stability phenomenon is observed in the offline setting (\ref{fig:dqn fully offline_space}, \ref{fig:dqn 0.3 perturbed_space} and \ref{fig:dqn last million_space}): the leading eigenvalue $\lambda_1$  rises to the quadratic threshold, after which it hovers around the quadratic threshold, albeit with some noise; when $\lambda_1$ reaches the quadratic divergence threshold, we observe more loss instability, but this does not always get reflected in the agent's reward. In online training, $\lambda_1$ excedees the quadratic divergence threshold. Here SGD with momentum was used.} 
    \label{fig:dqnalldatasets_space}
\end{figure}

In Figure~\ref{fig:dqnalldatasets_space} we present DQN results on Space Invaders with SGD with momentum which clearly show an edge of stability effect in every offline setting. In the online regime, the eigenvalues consistently rise above the quadratic threshold and it has no effect on the trend of the sharpness $(\lambda_1)$.

\begin{figure}[H]
    \centering
    \begin{subfigure}[b]{0.475\textwidth}
        \centering
        \includegraphics[width=\textwidth]{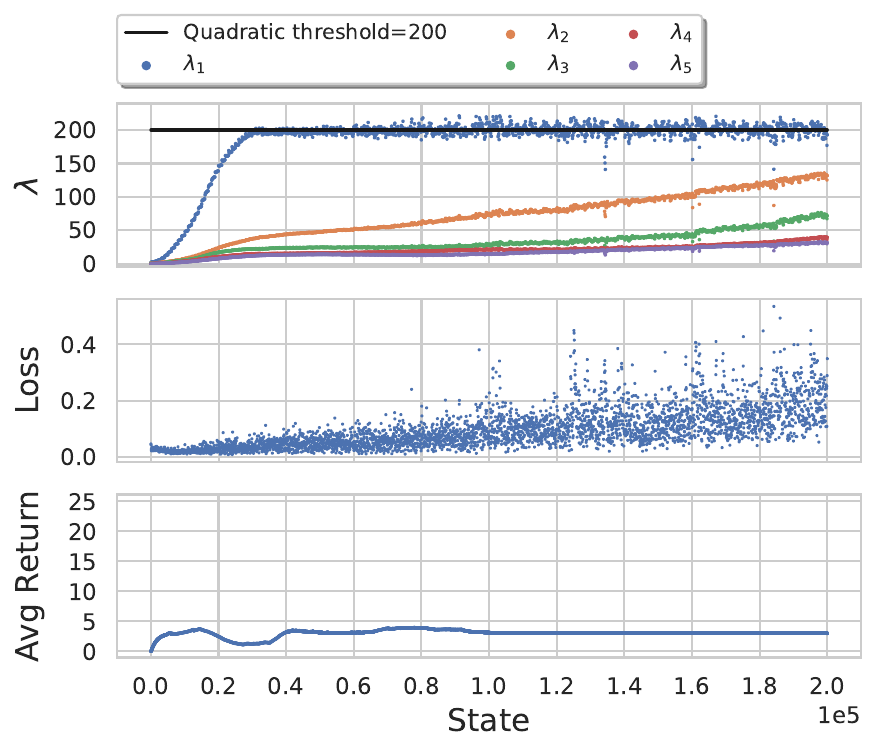}
        \caption[]%
        {\settingA.}    
        \label{fig:dqn fully offline_spacesgd}
    \end{subfigure}
    \hfill
    \begin{subfigure}[b]{0.475\textwidth}  
        \centering 
        \includegraphics[width=\textwidth]{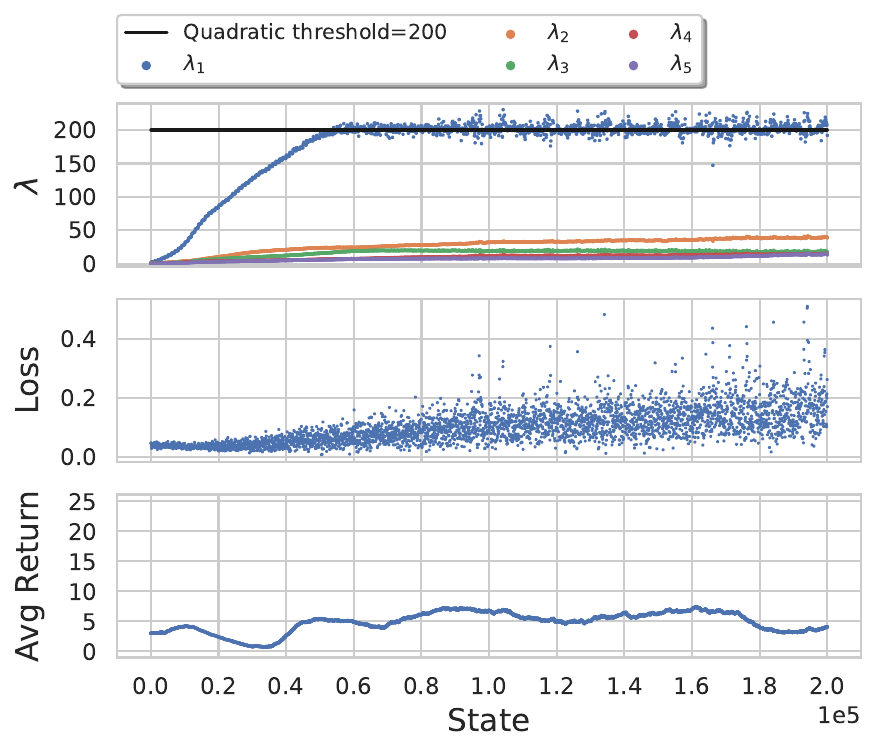}
        \caption[]%
        {\settingB.}    
        \label{fig:dqn 0.3 perturbed_spacesgd}
    \end{subfigure}
    \vskip\baselineskip
    \begin{subfigure}[b]{0.475\textwidth}   
        \centering 
        \includegraphics[width=\textwidth]{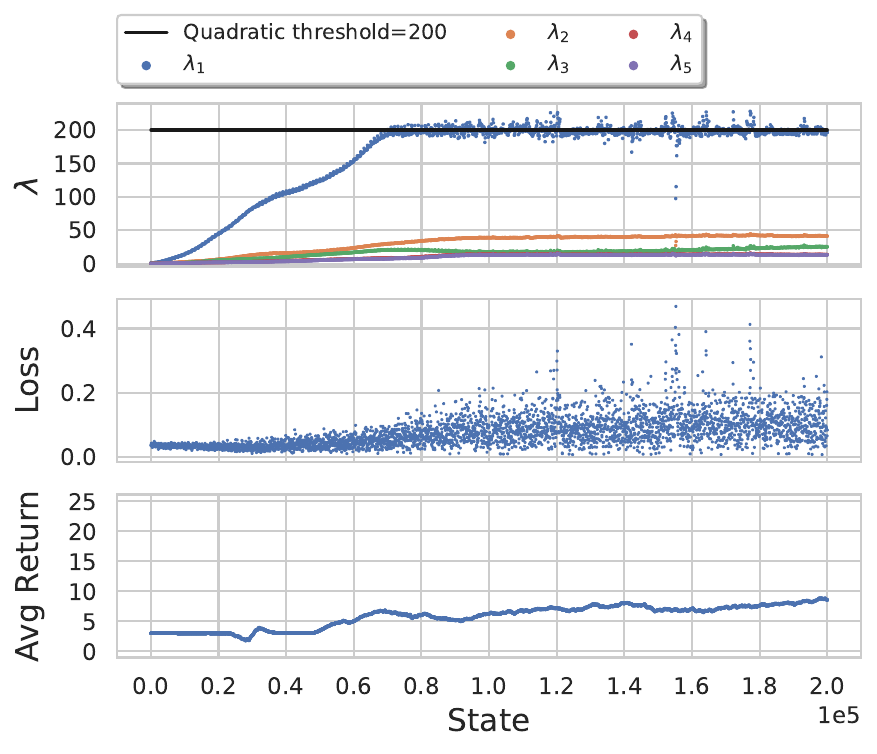}
        \caption[]%
        {\settingC.}    
        \label{fig:dqn last million_spacesgd}
    \end{subfigure}
    \hfill
    \begin{subfigure}[b]{0.475\textwidth}   
        \centering 
        \includegraphics[width=\textwidth]{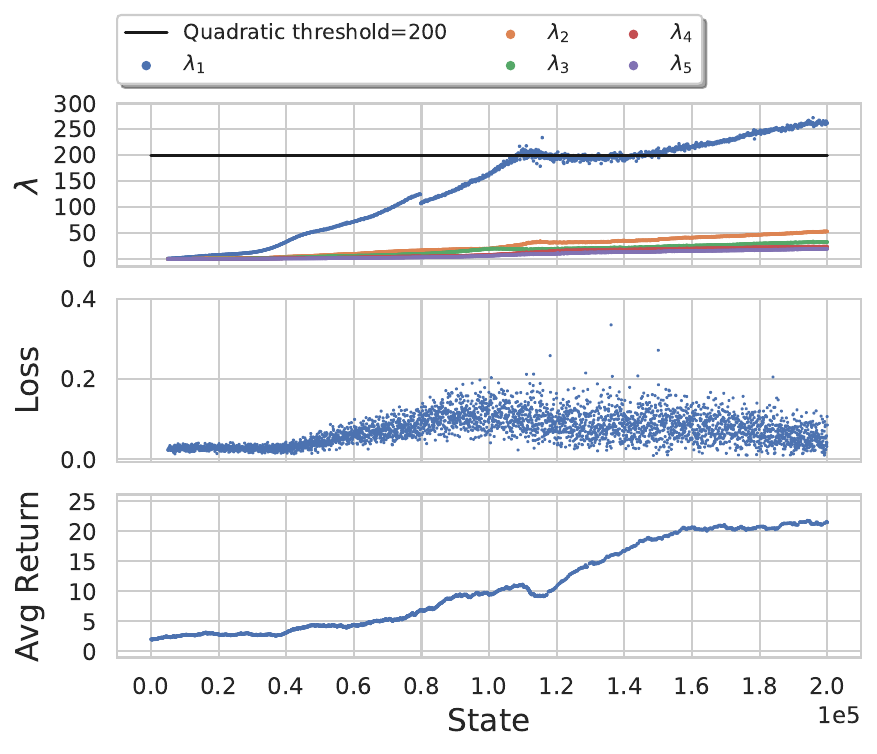}
        \caption[]%
        {\settingD.}    
        \label{fig:dqnonline_spacesgd}
    \end{subfigure}
    \caption[]
    {\textbf{DQN.} The edge of stability phenomenon is observed in the offline setting (\ref{fig:dqn fully offline_spacesgd}, \ref{fig:dqn 0.3 perturbed_spacesgd} and \ref{fig:dqn last million_spacesgd}): the leading eigenvalue $\lambda_1$  rises to the quadratic threshold, after which it hovers around the quadratic threshold, albeit with some noise; when $\lambda_1$ reaches the quadratic divergence threshold, we observe more loss instability, but this does not always get reflected in the agent's reward. In online training, $\lambda_1$ exhibits the edge of stability effect briefly but then it exceeds the quadratic divergence threshold. Here SGD without momentum was used.} 
    \label{fig:dqnalldatasets_spacesgd}
\end{figure}

In Figure~\ref{fig:dqnalldatasets_spacesgd} we present DQN results on Space Invaders with SGD without momentum which clearly show an edge of stability effect in every offline setting. In the online regime, there exists a trace of the edge of stability behaviour, however, later during training the principal eigenvalue consistently rises above the quadratic threshold.

\section{Experimental details}
\subsection{Neural architectures}

We used a similar neural network architecture for both DQN and C51. The network consists of 1 Convolutional Layer, followed by 1 Fully Connected (FC) Layer with and an Output Layer which depends on the algorithm. The Convolutional Layer has a kernel size of 3 and a stride of 1 and is configured differently based on the game due to different channel numbers. The rest of the details can be found in Table~\ref{app:network architecture}.

\begin{table}[H]
\centering

\caption{Parameter settings for the neural network architectures.}\label{app:network architecture}
\begin{tabular}{cccc}

\hline
\multicolumn{1}{l}{\textbf{Algorithm}} & \multicolumn{1}{l}{\textbf{Convolutional Layer width}} & \multicolumn{1}{l}{\textbf{FC Layer width}} & \multicolumn{1}{l}{\textbf{Output Layer width}} \\ \hline
DQN                                    & 16                                                     & 128                                         & $num\_actions$                                     \\ \hline
C51                                    & 16                                                     & 128                                         & $num\_actions \times num\_atoms$                         \\ \hline
\end{tabular}

\end{table}

\subsection{Algorithms}
\label{app:algo}
The pseudocode for the DQN algorithm is presented below\footnote{A detailed describtion can be found in \cite{mnih2013, mnihpaper2015}}:

\begin{algorithm}[hbt!]
\caption{Deep Q-learning with experience replay}\label{dqnalgo}

Initialize replay memory $D$ to capacity $N$\\
Initialize action-value function $Q$ with random weights $\theta$ \\
Initialize target action-value function $\hat{Q}$ with random weights $\hat{\theta} = \theta$\\
\For{episode=$1, \dots, M$}{
    Initialize sequence $s_1 = \{ x_1 \}$ and pre-processed state $\phi_1 = \phi (s_1)$
    \For{$t=0, 1, \dots, T$}{
        With probability $\epsilon$ select random action $a_t$ \\
        otherwise select $a_t = \argmax_a Q(\phi (s_{t}), a; \theta)$\\
        Execute action $a_t$ in emulator and observe reward $r_t$ and image $x_{t+1}$\\
        Set $s_{t+1} = s_t, a_t, x_{t+1}$ and preprocess $\phi_{t+1} = \phi (s_{t+1})$\\
        Store transition $\left ( \phi_{t}, a_t, r_t, \phi_{t+1} \right )$\\
        Sample random minibatch of transitions $\left ( \phi_{j}, a_j, r_j \phi_{j+1} \right )$\\
        Set $y_j = \left\{\begin{array}{lr}
        r_j, &  \text{if episode terminates at step $j+1$}\\
        r_j + \gamma \max_{a'} \hat{Q}(\phi_{j+1}, a'; \hat{\theta}), & \text{otherwise}\\
        \end{array} \right.$ \\
        Perform gradient step on $\mathcal{L} (y_j, Q(\phi_{j}, a_j; \theta))$ with respect to the network parameters $\theta$\\
        Every $C$ steps reset $\hat{Q} = Q$
    }
}
\textbf{Output}: Optimal $\pi \approx \pi_{*}$
\end{algorithm}

As an extension of DQN, \cite{c51} proposed to look at the entire value distribution dubbed $Z$ instead of considering expectations. Such a view permits the definition of distributional Bellman equations and operators\footnote{The Mathematics behind Categorical DQN is expanded in \cite{c51}.}. $Z$ is described discretely by a number $N \in \mathbb{N}$ and $V_{MIN}, V_{MAX} \in \mathbb{R}$, and whose support is the set of atoms $\left \{ z_i = V_{MIN} + i \Delta z : 0 \leq i < N \right \}$ with $\Delta z = \frac{V_{MAX} - V_{MIN}}{N - 1}$. These atoms represent the "canonical returns" of the distribution and each has probability given by a parametric model $\theta : \mathcal{S} \times \mathcal{A} \rightarrow \mathbb{R}^N$:
\begin{align}
    Z_{\theta} (s, a) = z_i \text{ with probability } p_i(s, a) = \frac{e^{\theta_i(s, a)}}{\sum_j e^{\theta_j(s, a)}}
\end{align}
\noindent The update is computed via $\hat{\mathcal{T}} Z_{\theta}$ where $\hat{\mathcal{T}}$ is an operator but a discrete distributional view poses a problem because because $Z_{\theta}$ and $\hat{\mathcal{T}} Z_{\theta}$ almost always have disjoint sets. To combat this issue, the update is reduced to multi-class classification by being projected onto the support of $Z_{\theta}$. Assume that $\pi$ is the greedy policy w.r.t $\mathbb{E} [Z_{\theta}]$. Given a tuple $(s, a, r, s', \gamma)$ the term $\hat{\mathcal{T}} z_j = r + \gamma z_j$ for each atom $z_j$. The probability $p_j(s', \pi(s'))$ is distributed to the immediate neighbours of $\hat{\mathcal{T}} z_j$ via a projection operator $\Phi$ whose $i^{th}$ component is given by\footnote{The quantity $\left [ \cdot \right ]_a^b$ bounds the argument in the interval $[a, b]$.}:
\begin{align}
    \left ( \Phi \hat{\mathcal{T}} Z_{\theta} (s, a) \right )_i = \sum_{j=0}^{N-1}  \left [ 1 - \frac{\left \lvert \left [ \hat{\mathcal{T}} z_j \right ]_{V_{MIN}}^{V_{MAX}} - z_i \right \rvert }{\Delta z} \right ]_0^1 p_j (s', \pi(s'))
\end{align}
\noindent In the end, as a DQN derivative, a policy network and a target network model $Z_{\theta}$ and $Z_{\hat{\theta}}$ (respecting the notation of Algorithm \ref{dqnalgo}) with the loss $\mathcal{L}$ given by the cross-entropy term of the KL Divergence:
\begin{align}
    D_{KL}\left ( \Phi \hat{\mathcal{T}} Z_{\hat{\theta}} (s, a) ||  Z_{\theta} (s, a) \right )
\end{align}
\noindent The routine of Categorical DQN is the same as before with the only expectation being the loss computation which is given by:
\begin{algorithm}[hbt!]
\caption{The Categorical algorithm \textbf{Input}: tuple $\left ( s_t, a_t, r_t, s_{t+1}, \gamma_t \in [0, 1] \right )$}\label{categoricaldqnalgo}

$\hat{Q}(\phi_{t+1}, a) \coloneqq \sum_i z_i p_i (\phi_{t+1}, a)$\\
$a^* \gets \argmax_a \hat{Q}(\phi_{t+1}, a)$\\
$m_i = 0, i \in \{ 0, 1, \dots, N-1 \}$\\
\For{$j = 0, 1, \dots, N-1$}{
    $\hat{\mathcal{T}} z_j \gets \left [ r_t + \gamma_t z_j \right ]_{V_{MIN}}^{V_{MAX}}$\\
    $b_j \gets (\hat{\mathcal{T}} z_j - V_{MIN}) / \Delta z$\\
    $l \gets \lfloor b_j \rfloor, u \gets \lceil b_j \rceil$\\
    $m_l \gets m_l + p_j (\phi_{t+1}, a^*)(u - b_j)$\\
    $m_u \gets m_u + p_j (\phi_{t+1}, a^*)(b_j - l)$
}

\textbf{Output}: $-\sum_i m_i \text{log } p_i (s_t, a_t)$ \Comment{Cross-entropy loss}
\end{algorithm}

\subsection{Offline RL reproduction details}
\label{app:offline_rl}

In this paper we examined three different offline RL replay buffers for Breakout on Minatar:

\begin{itemize}
    \item $10^6$ transitions that were obtained from the experience of a pre-trained agent with no action perturbation.
    \item $10^6$ transitions that were obtained from the experience of a pre-trained agent where during game-play 30\% of the actions taken were taken were random (instead of greedy actions being taken).
    \item $10^6$ transitions that were obtained from last $10^6$ transitions from the replay buffer of an agent that was trained with Adam, online.
\end{itemize}

\subsection{Optimistion details and how to replicate results}\label{app:replicateres}

We studied both the full-batch the mini-batch settings of GD and momentum for both algorithms. The mini-batch experiments always used a batch size of 512 and the full-batch experiments were performed on a sub-sample of $10^4$ transitions from the original replay buffers which consisted of $10^6$ transitions.

\begin{figure}[H]
    \centering
    \includegraphics[width=0.43\textwidth]{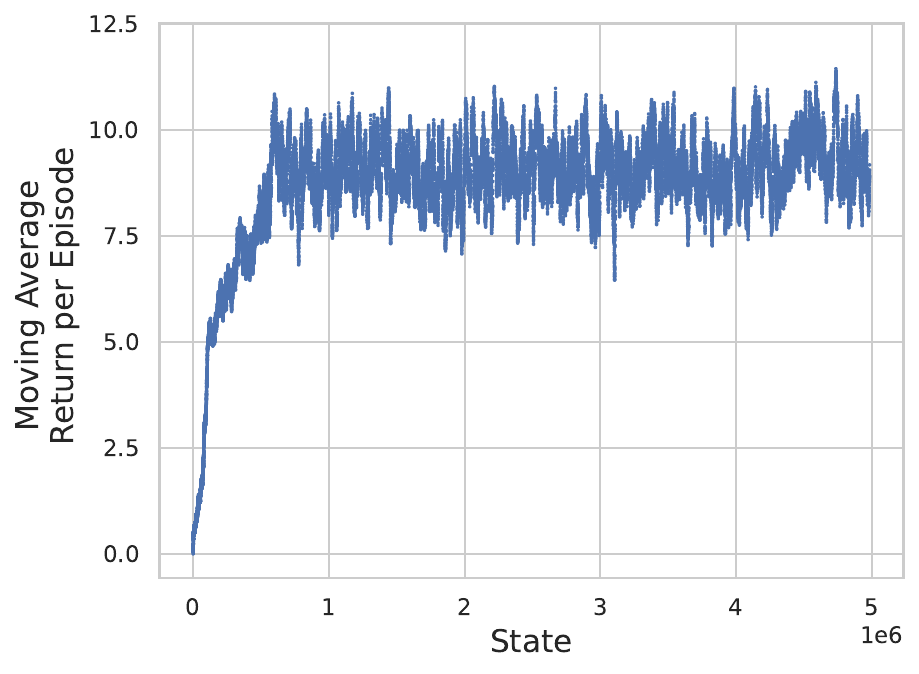}
    \caption{Adam results on Breakout showing the moving average return obtained by the agent during online training.}
    \label{fig:adambreakout}
\end{figure}

When experimenting with gradient descent, the learning rate was $0.01$ and when adding momentum the learning rate was $0.01$ and the momentum coefficient was $\beta = 0.8$. The initial $\gamma$ parameter to discount reward was $0.99$. The agent which generated the replay buffers was trained with Adam with a batch size of 64, learning rate of 0.00025, $\beta_1 = 0.9$, $\beta_2 = 0.999$ and $\epsilon = 10^{-8}$. Figure~\ref{fig:adambreakout} shows the return obtained for Breakout online with these settings. During offline training, the action executed by the agent is the action present in the replay buffer. During online training, the first $5000$ iterations are used to accumulate experiences in the replay buffer, after which the training of the agent starts. During the first $5000$ steps the agent is taking random actions. Afterwards, the actions are taken based on a decaying $\epsilon$-greedy policy where $\epsilon$ decreases linearly from $1.0$ to $0.1$ for $100000$ iterations. For C51, we used 51 atoms with $V_{MIN} = -10$ and $V_{MAX} = 10$.

The eigenvalues were logged at every 100 iterations. Whenever "Avg Return" was mentioned, that referred to a moving average of the return per episode. It was calculated based on the following routine: $avg\_return[i] = 0.99 * avg\_return[i-1] + 0.01 * return\_per\_episode[i]$ where $avg\_return[0] = return\_per\_episode[0]$.

The datasets for Breakout and Space Invaders used for experiments are available \href{https://github.com/riordan45/rl-edge-of-stability}{here}. In addition, we are able to provide datasets for performing similar experiments on Asterix, Freeway and Seaquest, the other games present in the Minatar testbed.

\end{document}